\documentclass[journal]{IEEEtran}
\usepackage{cite}

\def\etal{\textit{et al.}}

\ifCLASSINFOpdf
  \usepackage[pdftex]{graphicx}
  \graphicspath{{figures/}}
  \DeclareGraphicsExtensions{.pdf,.jpeg,.png,.jpg}
\else
\fi
\usepackage{multicol}
\usepackage{booktabs}
\usepackage{diagbox}

\usepackage{amsmath}
\ifCLASSOPTIONcompsoc
  \usepackage[caption=false,font=normalsize,labelfont=sf,textfont=sf]{subfig}
\else
  \usepackage[caption=false,font=footnotesize]{subfig}
\fi
\usepackage{url}

\begin{document}
%
\title{On the Evaluation and Real-World Usage Scenarios of Deep Vessel Segmentation for Retinography}
%
%
%

\author{Tim~Laibacher~and~Andr\'e~Anjos
\thanks{T. Laibacher was with Idiap Research Institute, Martigny 1920, Switzerland
(e-mail: tim.laibacher@gmail.com).}
\thanks{A. Anjos is with Idiap Research
Institute, Martigny 1920, Switzerland
(e-mail:  andre.anjos@idiap.ch).}
}

%
%

\markboth{Preprint}%
{Deep Vessel Segmentation for Retinography}
%



\maketitle

\begin{abstract}
    We identify and address three research gaps in the field of vessel
    segmentation for retinography.  The first focuses on the task of inference
    on high-resolution fundus images for which only a limited set of
    ground-truth data is publicly available. Notably, we highlight that simple
    rescaling and padding or cropping of lower resolution datasets is
    surprisingly effective.  We further explore the effectiveness of
    semi-supervised learning for better domain adaptation in this context.  Our
    results show competitive performance on a set of common public retina
    datasets, using a small and light-weight neural network.  For HRF, the only
    very high-resolution dataset currently available, we reach comparable, if
    not superior, state-of-the-art performance by solely relying on training
    images from lower-resolution datasets.  The second topic we address
    concerns the lack of standardisation in evaluation metrics.  We investigate
    the variability of the F1-score on the existing datasets and report results
    for recently published architectures.  Our evaluation show that most
    reported results are actually comparable to each other in performance.
    Finally, we address the issue of reproducibility, by open-sourcing the
    complete framework used to produce results shown here.
\end{abstract}

%

%
\IEEEpeerreviewmaketitle

\section{Introduction}
%
%
%
%
\IEEEPARstart{T}{he} accurate and automatic segmentation of retinal vasculature structure has several important applications in ophthalmology, such as in the diagnosis of diabetic retinopathy and wet age related macular degeneration.  This work identifies and addresses three gaps with high relevance for practical deployments.

The first concerns the \textit{un}availability of (annotated) high-resolution fundus images.  Owning to the popularity and ease of use of lower-resolution fundus datasets such as DRIVE~\cite{staal_ridge-based_2004} and STARE~\cite{hoover_locating_2000} for convolutional neural network training, the majority of previous works is still mostly focused on images that are magnitudes smaller than the fundus images which are today taken by clinics and practitioners around the world.  To the best of our knowledge, only the HRF dataset~\cite{budai_robust_2013} has a resolution that reaches, or comes close to the resolution taken by modern fundus cameras.
Since manual annotation by experts is time consuming and costly, we propose a set of methods to leverage existing public low-resolution datasets with annotated ground-truth vessel labels to train convolutional-neural networks that perform well on \textbf{unseen} high-resolution images.

In available literature, the performance evaluation of vessel segmentation models is often done by reporting the average F1-score across images in a (separate) test set.  Unfortunately, the way to calculate such average value is often omitted, leading to differences in reported values.  We note extrapolated standard deviations, a reasonable proxy for the average F1-score estimation uncertainty, is also left out in published work.  To close this gap, we outline two averaging methods for the F1-score: an opaque version, which is used in many publications in the field, and a second method that allows the natural extrapolation of the standard deviation to the reported average values.  The second averaging technique for the F1-score provides an interesting perspective into the interpretation of current state-of-the-art results.   We further our contribution by introducing a set of plots containing standard deviation bands, that confirm our analysis and provide additional insights into the robustness of models.

Finally we address the issue of reproducibility.  We found that the degree of reproducibility in previous work is rather unsatisfactory.  Because available public datasets do not share a common structure, programmatic access is particularly missing on existing contributions, which makes it very difficult to reproduce evaluation protocols reported in literature.  As a result, a large amount of engineering work is often necessary before actual research can take place.  To address this, we introduce the bob.ip.binseg package\footnote{\url{https://gitlab.idiap.ch/bob/bob.ip.binseg}} that integrates with the Bob framework~\cite{anjos_bob_2017}, can be easily extended, and allows for the reproduction of our experiments.

The remainder of this article is divided in 7 sections. Section~\ref{sec:review} describes related work using decoder/encoder deep neural network architectures (DNN).  Section~\ref{sec:baselines} introduces the standard training and evaluation pipeline, public datasets, evaluation metrics and baseline results for four DNN.  The proposed methods are detailed in Section~\ref{sec:methods} and the conducted experiments in Section~\ref{sec:experiments}.  Section~\ref{sec:evaluation} evaluates and compares the results against previous work, followed by Section~\ref{sec:conclusion} that concludes the article.

\section{Related Work}
\label{sec:review}

We split our review in 3 parts, matching each of our contributions in this work, namely: (i) A set of methods to leverage existing public low-resolution datasets to train convolutional neural networks that perform well on \textbf{unseen} high-resolution images; (ii) An analysis framework that provides insights on the uncertainty of reported average F1-scores and, finally, (iii) a reproducible set of methods that implement semantic segmentation of vascular structures in fundus images.

\subsection{High-Resolution Images}

In this section we describe related work that touches or focuses on the High-resolution Fundus Dataset (HRF, \cite{budai_robust_2013}).  We start by noting most state-of-the-art contributions are based on fully convolutional deep neural networks (FCN) composed of an encoding/decoding architecture for regressing vessel maps, with differences that can be summarized as follows:

\begin{enumerate}
    \item Small fully convolutional neural networks, trained on full-resolution images~\cite{laibacher_m2u_2018}.
    \item Large fully convolutional neural networks, trained on downsampled images~\cite{yan_joint_2018}.
    \item Patch-based training of fully convolutional neural networks with deformable convolutions~\cite{jin_dunet_2019}.
    \item Approaches with Generative Adversarial Networks ~\cite{goodfellow_gan_2014} (GANs) using downsampled images ~\cite{zhao_supervised_2019}.
\end{enumerate}

Due to it's proven record on various segmentation domains FCNs based on VGG16~\cite{simonyan_vgg_2015} as base encoder are employed in the majority of works such as  \cite{jin_dunet_2019,meyer_deep_2017,yan_joint_2018,zhao_supervised_2019}. The M2U-Net introduced in \cite{laibacher_m2u_2018} adopts a structure similar to the U-Net~\cite{ronneberger_u-net_2015} in ~\cite{yan_joint_2018}, relies however on MobileNetV2 in the encoder part and proposes light-weight inverted contracting residuals blocks in the decoder part.  Similarly, the DUNet by Jin~\etal~\cite{jin_dunet_2019} adopts the \emph{U} structure but uses Deformable Convolutions~\cite{dai_deformable_2016} in parts of the network.

The different methods come with a computational/pipeline complexity and time \textit{versus} segmentation quality trade-off as indicated in Table~\ref{tab:hrfinto}.  The deformable-convolution in DUNet, while light in terms of parameter count, impose a great reduction in inference speed as reported by the authors (47.7s vs 9.7s for an 999 x 960 image).  Additionally, patch-based inference pipelines are estimated to be slower in inference than methods that utilize full resolution~\cite{laibacher_m2u_2018} or downsampled images~\cite{yan_joint_2018}. The GAN based-approach by Zhao~\etal~\cite{zhao_supervised_2019} requires a two-step training procedure.  In the first step a synthesized target dataset is constructed using a modified GAN. In the second step, DRIU~\cite{maninis_deep_2016} is trained on the created synthesized images.

\begingroup
\setlength\tabcolsep{4pt} 
\begin{table}[htb]
\scriptsize
\caption{
Summary of previous works on the high-resolution HRF dataset when compared with respect to their reported performance and estimated number of trainable parameters.
}
\label{tab:hrfinto}
\begin{center}
\begin{tabular}{@{}ccccccc@{}}
\toprule
                                    
Method                                                          & Year & F1              & GAN   & Patch-based & Parameter &              \\ \hline
Orlando et al.~\cite{orlando_discriminatively_2017}*            & 2017 & 0.7158          & No    & No          & -               \\
Yan et al.~\cite{yan_joint_2018}*                               & 2018 & 0.7212          & No    & No          & ~25.85M            \\
Laibacher et al.~\cite{laibacher_m2u_2018}*                     & 2018 & 0.7814          & No    & No          & 0.55M                \\ 
Jin et al.~\cite{jin_dunet_2019}*                               & 2019 & \textbf{0.7988} & No    & Yes         & 0.88M                \\
Zhao et al.~\cite{zhao_supervised_2019}                         & 2019 & 0.7659          & Yes   & No          & 14.94M               \\        
DRIU~\cite{maninis_deep_2016} (our impl.)*                      & 2019 & 0.7865          & No    & No          & 14.94M               \\     
\multicolumn{7}{l}{*Same train-test split}  \\ \bottomrule
\end{tabular}
\end{center}
\end{table}
\endgroup

Taking into account these trade-offs, we argue that the proposed methods remain largely comparable in performance and minor reported improvements do not represent significant breakthroughs in semantic vessel segmentation for fundus images.

\subsection{Evaluation Metrics}
\label{sec:intrometrics}

A coherent and transparent standard for evaluation metrics is necessary to allow for a fair comparison of methods.  In the field of vessel segmentation however, reported metrics frequently differ both in terms of kind and quantity.  While the F1-score has emerged as one of the dominant metrics, exact details on it's calculation are often not clearly stated.

FCNs commonly output probability maps, attaching a vessel probability score (typically, a floating-point number in the range $[0, 1]$) to each pixel in the image.  Since the ground-truth labels are binary, the probability map has to be thresholded.  In ~\cite{maninis_deep_2016,laibacher_m2u_2018,jin_dunet_2019,zhao_supervised_2019} metrics on the test-set are evaluated at all thresholds, in steps of 0.01.  Metrics at the optimal test set threshold (\textit{a posteriori}) are reported in ~\cite{fraz_ensemble_2012,marin_new_2011,meyer_deep_2017}. Li~\etal~\cite{li_cross-modality_2016} utilize the threshold determined on the training set (\textit{a priori}), a scheme also adopted by Yan~\etal~\cite{yan_joint_2018}, and that we find to be less biased.

The differences in evaluation metrics are identified as the first barrier for fair comparison and evaluation of competing methods.  The second is the difference in training and test splits.  Out of the five considered datasets, only DRIVE~\cite{staal_ridge-based_2004} defines a train-test split. The remaining datasets leave it to the author to define an appropriate split.  As will become evident in Section~\ref{sec:experiments}, while in some cases a dominant split has emerged in the literature, in other cases the utilized splits differ considerably, further hindering fair comparisons.

In this work both barriers are addressed, by clearly describing metric calculations and train-test splits in the hopes of inspiring future work to adopt a similar approach.  We further push the reproducibility envelop by providing APIs for accessing such splits in a programmatic way.

\subsection{Reproducibility}

Machine learning experiments are becoming increasingly complex, making it harder to fully reproduce proposed systems reliably~\cite{anjos_bob_2017}.  This problem is especially pronounced in niche computer vision fields like vessel segmentation, that receive less attention compared to popular image-classification or object detection tasks in which reproducible work is found more frequently.  The maskrcnn\_benchmark~\cite{massa_mrcnn_2018} for example, since it's introduction, saw several independent contributions and publications building on top of its framework~\cite{tian_fcos_2019,fu_retinamask_2019}.
Besides being good practice, reproducible research has shown to be beneficial to the impact of publications~\cite{vandewalle_reproducible_2009}. Vandewalle~\etal~\cite{vandewalle_reproducible_2009} distinguish six degrees of reproducibility, ranging from easily reproducible (5) to not reproducible (0). We refer to the paper for the exact definitions.

We found that existing works do either not provide any source-code at all ~\cite{fraz_ensemble_2012,li_cross-modality_2016, liskowski_segmenting_2016, orlando_discriminatively_2017}, estimated to require extreme effort to reproduce (2), provide source-code that lack documentation and instructions on how to setup the training environment and datasets ~\cite{maninis_deep_2016, zhao_supervised_2019,jin_dunet_2019} or provide only parts of the training pipeline~\cite{yan_joint_2018} and therefore require considerable effort to reproduce (3).

This highlights the need for full reproducible work in this domain, which we address in the form of a comprehensive software-package (bob.ip.binseg) that is attached to this publication.

\renewcommand{\arraystretch}{1.3}


\begin{table}[ht]
\caption{Overview of retina vessel segmentation datasets.}
\label{tab:datasets}
\begin{center}
\begin{tabular}{@{}cccccc@{}}
\toprule
Dataset    & H x W       & Imgs.   & Train & Test & Reference                 \\ \midrule
DRIVE      & 584 x 565   & 40      & 20    & 20   & \cite{staal_ridge-based_2004}   \\
STARE      & 605 x 700   & 20      & 10    & 10   & \cite{maninis_deep_2016}        \\
CHASE\_DB1 & 960 x 999   & 28      & 8     & 20   & \cite{fraz_ensemble_2012,laibacher_m2u_2018}  \\
IOSTAR     & 1024 x 1024 & 30      & 20    & 10   & \cite{meyer_deep_2017}             \\
HRF        & 2336 x 3504 & 45      & 15    & 30   & \cite{orlando_discriminatively_2017,laibacher_m2u_2018,jin_dunet_2019} \\ \bottomrule
\end{tabular}
\end{center}
\end{table}

\section{Baseline Benchmarks}
\label{sec:baselines}
This section sets the stage for the remaining paper by illustrating the training-evaluation pipeline, public datasets and metrics.  Additionally, we establish baselines for a set of popular FCNs for vessel segmentation.

\subsection{Pipeline}
\begin{figure}[htb]
    \centering
    \includegraphics[width=0.49\textwidth]{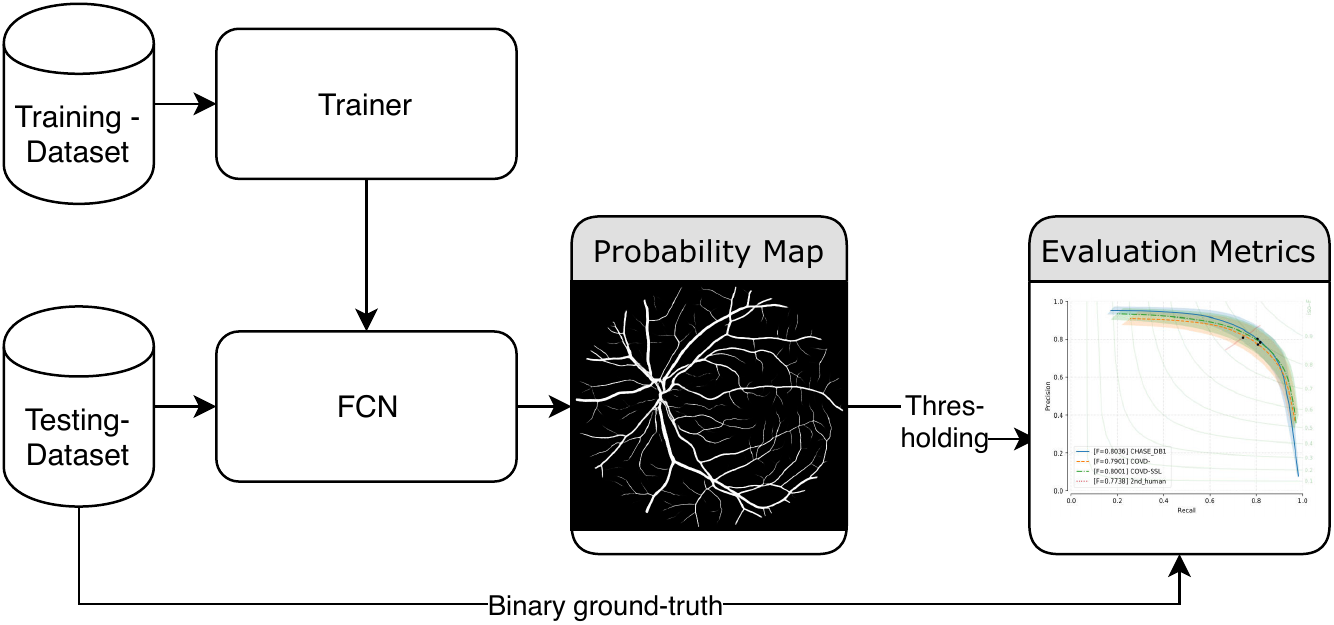}
    \caption{Typical FCN training/testing pipeline on vessel segmentation datasets}    \label{fig:pipeline}
\end{figure}

A common FCN training/testing pipeline is illustrated in Figure~\ref{fig:pipeline}. During training, the training-split or dataset is fed into the FCN, with images and ground-truth pairs often being augmented using changes in contrast, brightness and color and via rotation/flipping. In the testing-phase, the fundus images from the test-split or dataset are fed through the trained network, generating a probability map as an output. Since the ground-truth labels are binary, thresholding takes place before the evaluation metrics are calculated.

{}

\subsection{Datasets}
The five most commonly used publicly available datasets are DRIVE~\cite{staal_ridge-based_2004}, STARE~\cite{hoover_locating_2000}, CHASE\_DB1~\cite{owen_measuring_2009}, HRF~\cite{budai_robust_2013} and IOSTAR~\cite{abbasi_iostar_2015} with the order being indicative of their appearance in the literature.  As mentioned, the majority of datasets does not have a predefined train-test split and previous works have not always converged to a dominant split and thereby limiting the scope for a fair comparison. The splits adapted in this work are as follows:

For DRIVE, we use the train-test split as proposed by the authors of the dataset. For STARE, we follow  Maninis~\etal~\cite{maninis_deep_2016} and Zhao~\etal~\cite{zhao_supervised_2019} with a 10/10 split. The split adopted for CHASE\_DB1 was first proposed by Fraz~\etal~\cite{fraz_ensemble_2012}, which uses the first 8 images for training and the last 20 for testing. For HRF, we adopt the split as proposed by Orlando~\etal~\cite{orlando_discriminatively_2017}, and adapted in \cite{laibacher_m2u_2018} and \cite{jin_dunet_2019}, whereby the first five images of each category (healthy, diabetic retinopathy and glaucoma) are used for training and the remaining 30 for testing.  For IOSTAR, we select the 20/10 split introduced by Meyer~\etal~\cite{meyer_deep_2017}.  Table~\ref{tab:datasets} provides an compact overview of dataset sizes, resolutions, splits and references.

\subsection{Metrics}
In addition to different thresholding methods as highlighted in Section~\ref{sec:intrometrics}, the calculation of F1-scores can also differ.
The F1-score is derived from precision (Pr) and recall (Re) values, which in turn are derived from true positives (TP), false negatives (FN)  and false positives (FP), calculated for each predicted vessel map/ground-truth pair. Other common metrics include accuracy (Acc) and specificity (Sp):

\begin{equation}
    Pr = \frac{TP}{TP+FP}
\end{equation}
\begin{equation}
    Re = \frac{TP}{TP+FN}
\end{equation}
\begin{equation}
    Sp = \frac{TN}{FP+FN}
\end{equation}
\begin{equation}
    Acc = \frac{TP+TN}{TP+FP+FN+TN}
\end{equation}
\begin{equation}
    F1 = \frac{2*(Pr*Re)}{Pr+Re}
\end{equation}

The average F1-Score for all test-images can either be calculated on a micro level or on a macro level.  On the micro level, F1-scores for each test-images $i$ are averaged:

\begin{equation}
    \overline{F1_{micro}} = \frac{1}{n}\sum_{i=1}^n F1_i
\end{equation}

On the macro level, the F1-score is calculated based on average precision and recall:

\begin{equation}
    \overline{F1_{macro}} = \frac{2*(\overline{Pr}*\overline{Re})}{\overline{Pr}+\overline{Re}}
    \label{eq:f1macro}
\end{equation}

While previous published work uses the later, which typically leads to slightly higher scores, the former calculation method allows for additional insights on the variability of the model's performance since F1-score standard deviations can be included.

\subsection{FCN and Results}
In order to establish baseline results, the following four popular FCNs used for retinal vessel segmentation are evaluated on the aforementioned datasets and splits (number of trainable network parameters in brackets): DRIU (14.94M)~\cite{maninis_deep_2016}, HED (14.73M)~\cite{xie_holistically-nested_2015}, M2U-Net (0.55M)~\cite{laibacher_m2u_2018} and U-Net (25.85M)~\cite{ronneberger_u-net_2015}.  The results are shown in Table ~\ref{tab:baselines}.

The fact that the considerable smaller M2U-Net's performance almost reaches the performance of larger models like DRIU and U-Net, especially for the high resolution datasets HRF, hints at overparametrization of the pretrained ImageNet~\cite{imagenet_cvpr09} part of those models. Similar observations were made by Raghu \etal~\cite{raghu_transfusion_2019} on the RETINA~\cite{gulshan_development_2016} classification dataset.

Out of the four models, we picked DRIU and M2U-Net to apply our methods to.  M2U-Net was chosen because of the extreme compactness, and DRIU since it came very close to the heavy U-Net performance, while requiring less parameters.  We found that by saving in trainable parameters allows us to train networks to perform in high-resolution images using larger batch sizes, and that this consistently improves the overall classifier performance.

\begingroup
\setlength\tabcolsep{4pt} 
\begin{table}[ht]
\caption{Baseline benchmark results, models are trained and tested on the same dataset using the splits as indicated in Table~\ref{tab:datasets}}
\label{tab:baselines}
\begin{center}
\begin{tabular}{@{}ccccc@{}}
\toprule
$F1_{micro}$ (std)     & DRIU          & HED           & M2U-Net       & U-Net         \\ \midrule
CHASEDB1     & 0.810 (0.021) & 0.810 (0.022) & 0.802 (0.019) & \textbf{0.812} (0.020) \\
DRIVE        & 0.820 (0.014) & 0.817 (0.013) & 0.803 (0.014) & \textbf{0.822} (0.015) \\
HRF          & 0.783 (0.055) & 0.783 (0.058) & 0.780 (0.057) & \textbf{0.788} (0.051) \\
IOSTAR & \textbf{0.825} (0.020) & \textbf{0.825} (0.020) & 0.817 (0.020) & 0.818 (0.019) \\
STARE        & 0.827 (0.037) & 0.823 (0.037) & 0.815 (0.041) & \textbf{0.829} (0.042) \\ \bottomrule
\end{tabular}
\end{center}
\end{table}
\endgroup

\section{Methods}
\label{sec:methods}

In this section we describe two approaches for vessel segmentation for high-resolution images:  We first outline a proposed combined dataset, followed by our rescaling, padding and cropping scheme. Lastly we describe our approach to semi-supervised learning.  Here we use a simple scheme whereby three guesses of unlabeled images, created by the network during training, are averaged and incorporated in a combined loss function by a weighting factor.  Finally, we describe other implementation details and hyper-parameters.

\subsection{Combined Vessel Dataset}
To increase the robustness of the trained models, we propose training on a combined vessel dataset (COVD), consisting of the aforementioned five publicly available datasets.  Whenever we exclude the target dataset we use for testing from training, we indicate it by a $-$ sign. E.g. COVD$-$ tested on target dataset HRF, means we include all datasets for training, except HRF itself. Similarly, COVD$-$ evaluated on the target dataset CHASE\_DB1 means we include all datasets for training, except CHASE\_DB1 itself.  In this way we emulate more realistic use-cases where there often is no ground-truth data available.

In cases where we use semi-supervised learning we utilize the training images but \emph{not} the ground-truth data of the target dataset. E.g. For COVD$-$SSL evaluated on HRF, we utilize all datasets except HRF for training with ground-truth pairs and for SSL we use only the training images of the HRF training set.

\subsection{Rescaling, Cropping, Padding}
One challenge that arises when training and testing datasets with different resolutions, is how to perform rescaling. For single source-to-target dataset resizing, a common approach is to downscale the higher-resolution target dataset (e.g. HRF) to the resolution of the lower-resolution source dataset (e.g. DRIVE), train the network and during inference upsample the generated probability maps again to the target dataset resolution. See for example \cite{zhao_supervised_2019,yan_joint_2018}.

\begin{figure*}[!t]
\centering
\includegraphics[width=0.32\textwidth]{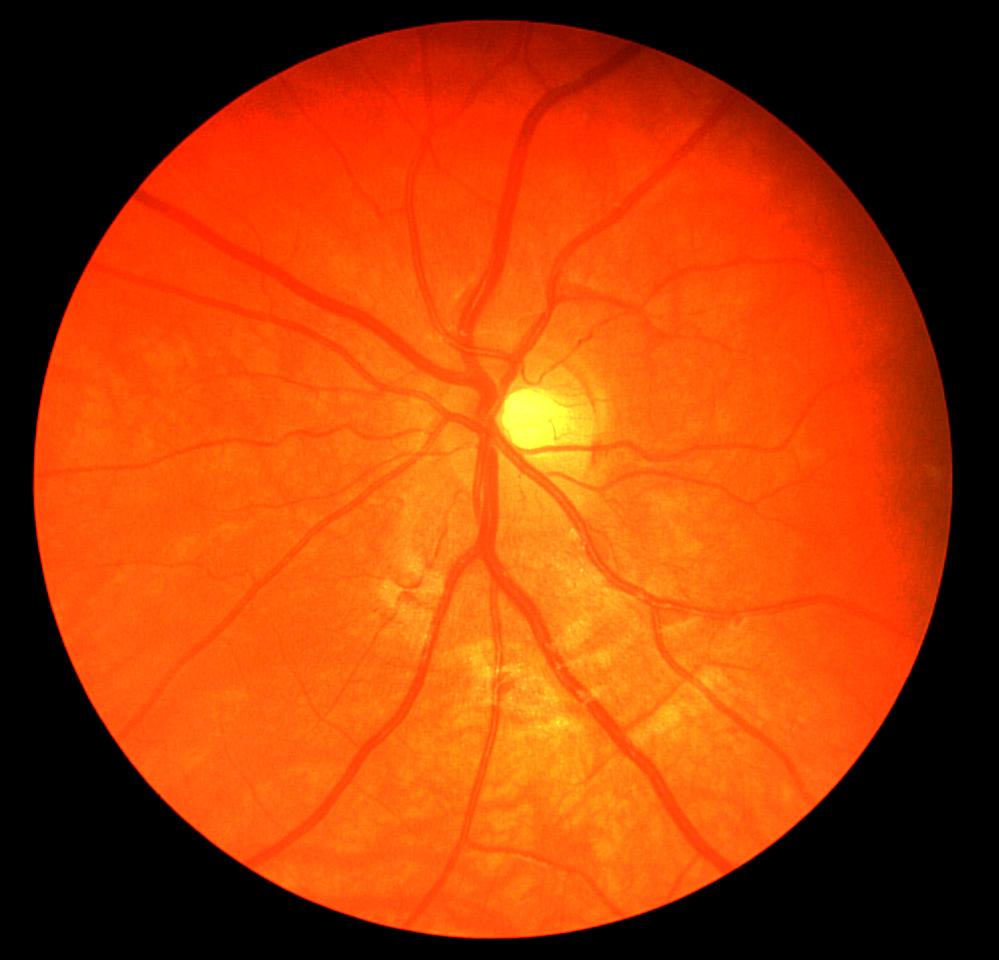}%
\quad%
\includegraphics[width=0.32\textwidth]{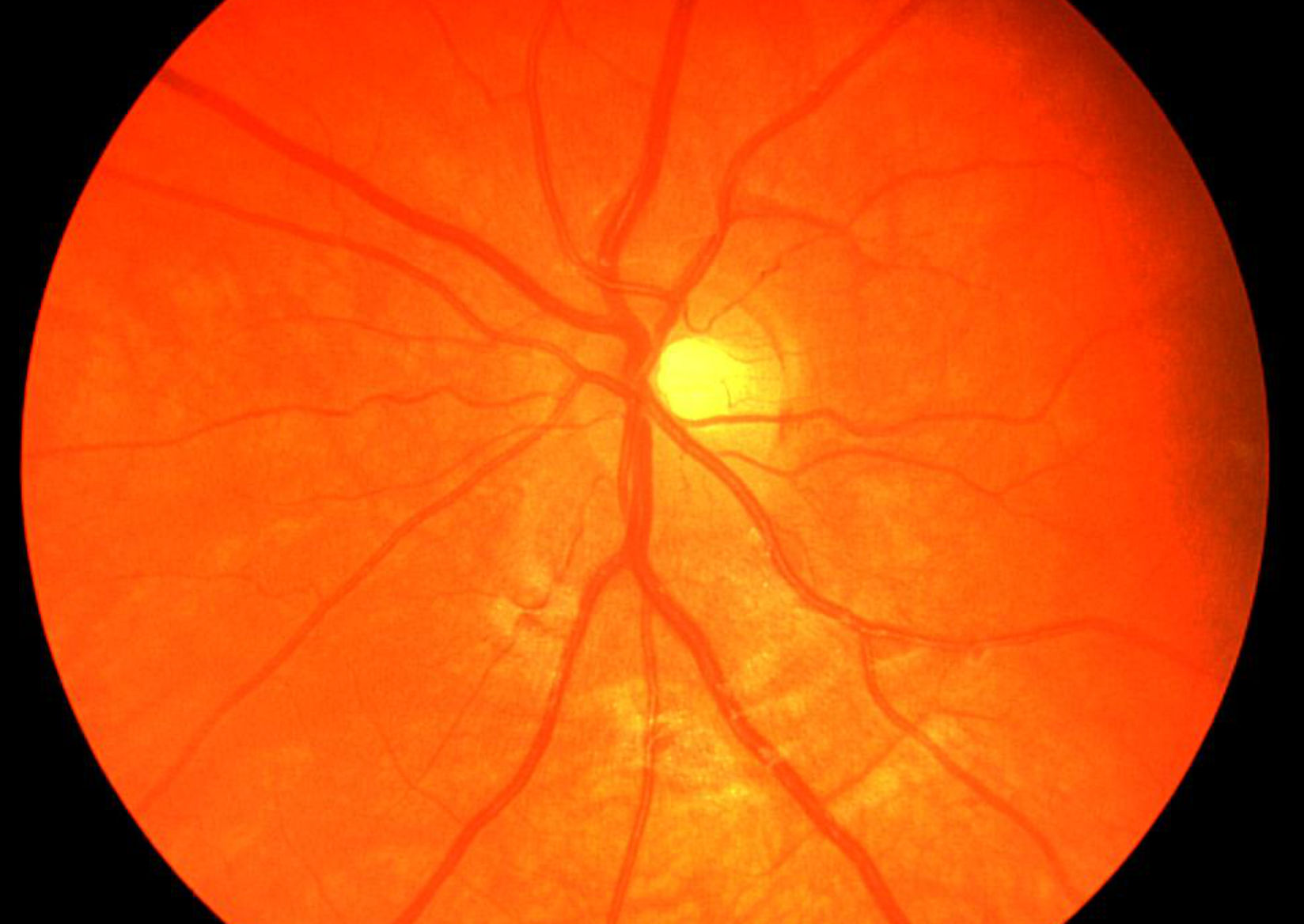}%
\quad%
\includegraphics[width=0.32\textwidth]{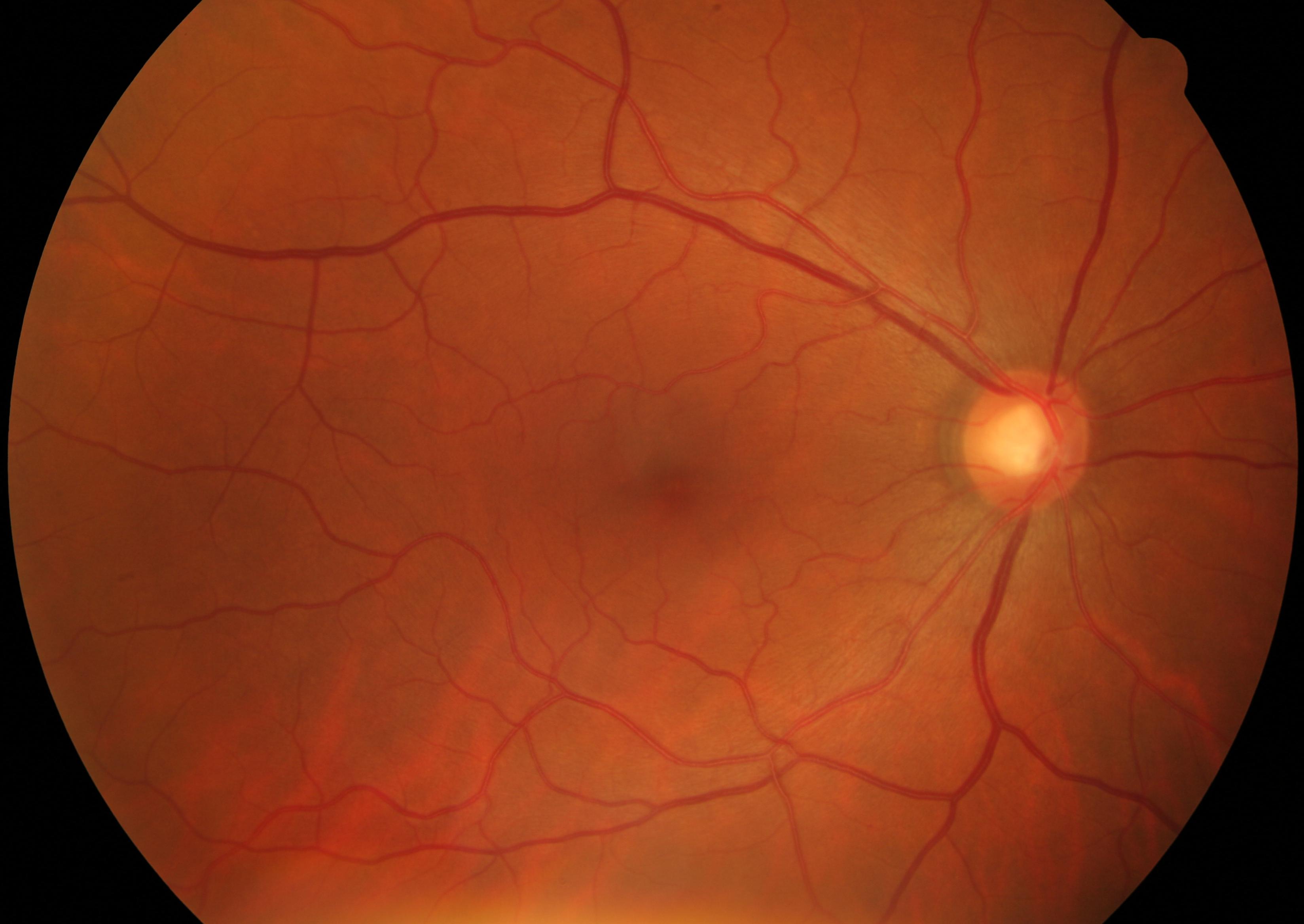}
\caption{Illustration of upscaling and cropping from source dataset CHASE\_DB1
to target dataset HRF. From left to right: CHASE\_DB1 image; Upscaled and
cropped to HRF resolution; and HRF image for comparison.}
\label{fig:rescale}
\end{figure*}

In contrast to this approach, we propose a scheme where the source dataset is resized, cropped and padded so that it has the resolution and spatial composition of the target dataset. The model is then trained on this modified source-dataset. This is best illustrated by an example:

Treating HRF as the target dataset, and CHASE\_DB1 as the source dataset, we first perform a crop on the latter followed by a resize operation as depicted in Figure~\ref{fig:rescale}.

Similar operations are conducted for each source-to-target combination for the remaining datasets.  We refer to the bob.ip.binseg package documentation for the exact operations performed for each case.

\subsection{Semi-Supervised Learning}
Semi-Supervised Learning (SSL)~\cite{chapelle_ssl_2010} has seen increased interest in the image classification domain, with recent works including~\cite{berthold_mixmatch_2019,olivier_contrastivessl_2019}.
In this work we adopt the approach of Berthold~\etal ~\cite{berthold_mixmatch_2019} of using unlabeled examples and labeled examples in separate loss terms that are combined by a weighting factor. Given a batch of unlabeled examples $\mathcal{U}$ from the target dataset, for each unlabeled image $u_b$ in the batch, three guessed probability vessel labels are generated via a forward pass through the model using the unlabeled image, a horizontally flipped version of it and a vertical flipping version of it which are averaged to form $\hat{g}_b$:
\begin{multline}
    g_b = \frac{1}{3} \left[ p_{model}(y|u_b) + p_{model}(y_{hflip}|u_{bhflip}) \right . \\
    \left . + p_{model}(y_{vflip}|u_{bvflip}) \right]
\end{multline}

and then used in the SSL-Loss covered in the following section.

\subsubsection{Loss Functions}
For standard supervised-learning without SSL we utilize the Jaccard Loss~\cite{iglovikov_ternausnetv2_2018} that is a combination of Binary Cross-Entropy loss and the Jaccard Score weighted by a factor $\alpha$:

\begin{equation}
    \label{eq:LJBCE}
    \mathcal{L}_{JBCE} = \alpha \mathcal{L}_{BCE} - (1 - \alpha)(1-J)
\end{equation}
We adopt $\alpha=0.3$ as suggested by Iglovikov~\etal~\cite{iglovikov_ternausnetv2_2018}.

The Binary Cross-Entropy loss, where $p_{model}(y_i|x_i)$ are values corresponding to predicted probability of a pixel $i$ belonging to the vessel class and $y_i$ is the ground-truth binary value, forms the first part of the equation and is defined as:
\begin{multline}
    \label{eq:BCE}
    \mathcal{L}_{BCE} = - \frac{1}{n} \sum^n_{i=1}(y_i log(p_{model}(y_i|x_i) + \\ (1-y_i) log(1-p_{model}(y_i|x_i)) )
\end{multline}

An adaptation of the Jaccard coefficient for continuous pixel-wise probabilities forms the second part of the combined loss function:
\begin{equation}
    \mathcal{J} = \frac{1}{n} \sum^n_{i=1} \frac{y_i*p_{model}(y_i|x_i)}{y_i+\hat{y}_i - y_i*p_{model}(y_i|x_i)}
\end{equation}
We note that the Jaccard coefficient has a monotonically increasing relation with the F1-Score (also know as Dice coefficient)~\cite{pont-tuset_supervised_2016}, so it can act as an appropriate loss-function even though our actual evaluation metrics is the F1-Score.

During SSL training, we further combine Equation \ref{eq:LJBCE} for labeled examples $\mathcal{X} = (x_b,y_b); b \in (1,...,B)$ and Equation \ref{eq:BCE} for unlabeled examples with guessed labels $\mathcal{U'}=(u_b,g_b); b \in (1,...,B)$, weighted by $\lambda$:
\begin{multline}
    \mathcal{L}_{ssl}(X,U') = \frac{1}{|\mathcal{X}|}\sum_{x,y \in\mathcal{X}}\mathcal{L}_{JBCE}(\mathcal{X}) + \\ \frac{1}{|\mathcal{U}|}\sum_{u,g \in\mathcal{U'}}\lambda*\mathcal{L}_{BCE}(\mathcal{U'})
\end{multline}


Instead of using a constant $\lambda$, we use a quadratic ramp-up schedule, with the intuition behind that the further we are in the training process the better the semi-supervised predictions should get.

\subsection{Advanced Precision vs. Recall Plot}
\label{sec:advprecisionrecall}

In addition to the $F_{micro}$ score with standard deviations, we here introduce an extended precision vs. recall plot that plots $\overline{Pr}$ and $\overline{Re}$ at every threshold together with the standard deviation in both precision and recall. In addition iso-F curves are plotted in light green and the point along the curve with the highest F1-score is highlighted in black. In this case, in order to have a consistent representation within the plot, the F1-score is macro averaged (Equation~\ref{eq:f1macro}). This setup allows for an easy visual comparison of model performance and their variability across test-images.

\begin{figure}[ht]
    \centering
    \includegraphics[width=.45\textwidth]{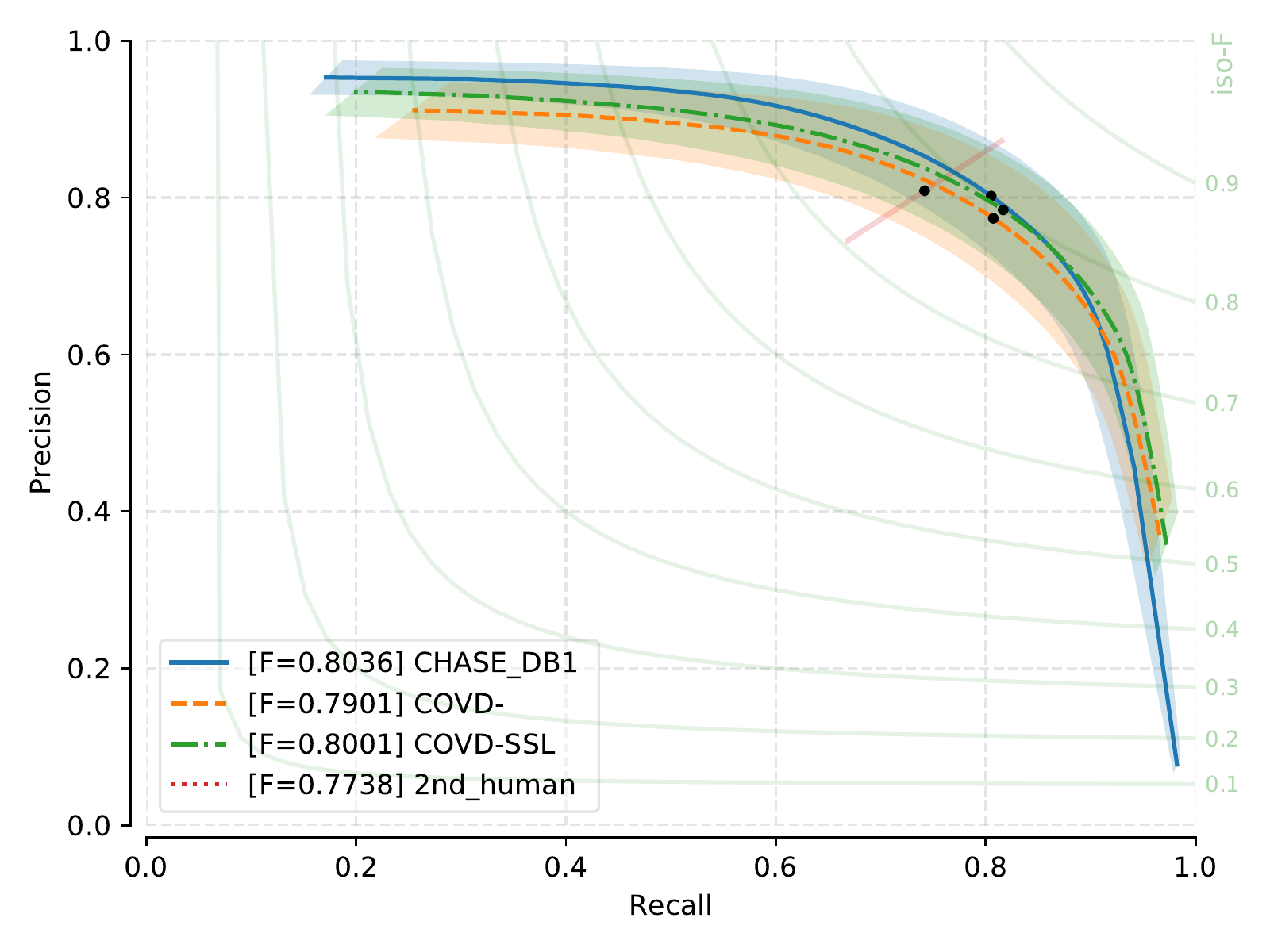}
    \caption{M2U-Net precision vs. recall curves on CHASE\_DB1. Bands in light colors depict standard deviations.}
    \label{fig:chasedb1_pr_1}
    \end{figure}

E.g. in Figure~\ref{fig:chasedb1_pr_1} it can be observed that the variability across CHASE\_DB1 annotations made by the 2nd human is higher than that of our models since their standard deviation bands are narrower compared to the standard deviation of the 2nd human annotator depicted by a single line in light red. Put simply, the models make more consistent predictions across test images compared to the second annotator.  It is interesting to observe that close to the highest F1-score for each model (the value typically reported in publications), the performance of all systems is quite comparable.  The region delimited by the upper and lower bounds defined by the standard deviations for any given system encompass all other systems.

\subsection{Implementation Details}
\label{sec:impldetails}
The following details apply to both, "normal" training and SSL training. We deploy AdaBound~\cite{luo_adaptive_2019} as an optimizer, using the default parameters as suggested in the original paper with a learning rate of 0.001. During the training-phase we apply the following random augmentations: horizontal flipping, vertical flipping, rotation and changes in brightness, contrast, saturation and hue. Training is conducted for 1000 epochs with a reduced learning rate of 0.0001 after 900 epochs. For all datasets except HRF, we use the original resolution and perform necessary cropping/padding so that the resolution is a multiple of 32, a requirement for the U-Net and M2U-Net variants. Since our training hardware setup did not have enough GPU memory for the training of the full resolution HRF images (and the upscaled COVD$-$ source datasets), we use half resolution images for training (1168 x 1648) but run inference on the full resolution images (2336 x 3296). We refer to the released bob-package and documentation for all details necessary to reproduce our results.

\section{Experiments}
\label{sec:experiments}
To evaluate the performance of our rescaling, padding and cropping scheme using COVD, we treated each of the datasets in Table~\ref{tab:datasets} in turn as the target dataset. Here we only report the results for M2U-Net and refer to additional material available on software package documentation for the results with DRIU.

On all tested datasets we found that training on COVD$-$ yields competitive results that come close to the performance of the baselines where the model was trained and \textbf{tested on the same dataset}. This is encouraging, given the large differences in illumination, contrast, color and resolution of the source datasets. For HRF, we can report performance improvements of almost 2 p.p. compared to the baseline.

Further applying SSL we gain additional improvements of around 1 p.p. for CHASE-DB1 and STARE, in the latter case now narrowly beating the baseline. For DRIVE we only found marginal improvements and worse performance for HRF and IOSTAR. We therefore fail to make conclusive statements about the viability of SSL for domain adaption and leave the investigation, mitigation and improvement of this method to future work.

Table~\ref{tab:ssl} summarizes the results for COVD$-$ and COVD$-$SSL. Additional precision vs. recall curves for all datasets are available as additional material in the documentation of the bob.ip.binseg package.
\begin{table}[ht]
\caption{$F1_{micro}$-score (std) for M2U-Nets on trained on Target, COVD$-$ and COVD$-$SSL. }
\label{tab:ssl}
\begin{center}
\begin{tabular}{l|c|cc@{}}
\toprule
\diagbox{Target}{Source} & Target                    & COVD$-$                   & COVD$-$SSL                   \\ \midrule
DRIVE                    & \textbf{0.803 (0.014)}    & 0.789 (0.018)           & 0.791 (0.014)              \\
STARE                    & 0.815 (0.041)             & 0.812 (0.046)           & \textbf{0.820 (0.044)}     \\
CHASEDB1                 & \textbf{0.802 (0.019)}    & 0.788 (0.024)           & 0.799 (0.026)              \\
HRF                      & 0.780 (0.057)             & \textbf{0.802 (0.045)}  & 0.797 (0.044)              \\
IOSTAR                   & \textbf{0.817 (0.020)}    & 0.793 (0.015)           & 0.785 (0.018)              \\ \bottomrule
\end{tabular}
\end{center}
\end{table}

\section{Evaluation}
\label{sec:evaluation}
\begin{figure*}[!t]
\centering
    \begin{minipage}[h]{0.02\textwidth}
    \rotatebox{90}{HRF}
    \end{minipage}
    \begin{minipage}[h]{0.32\textwidth}
        \includegraphics[width=\textwidth]{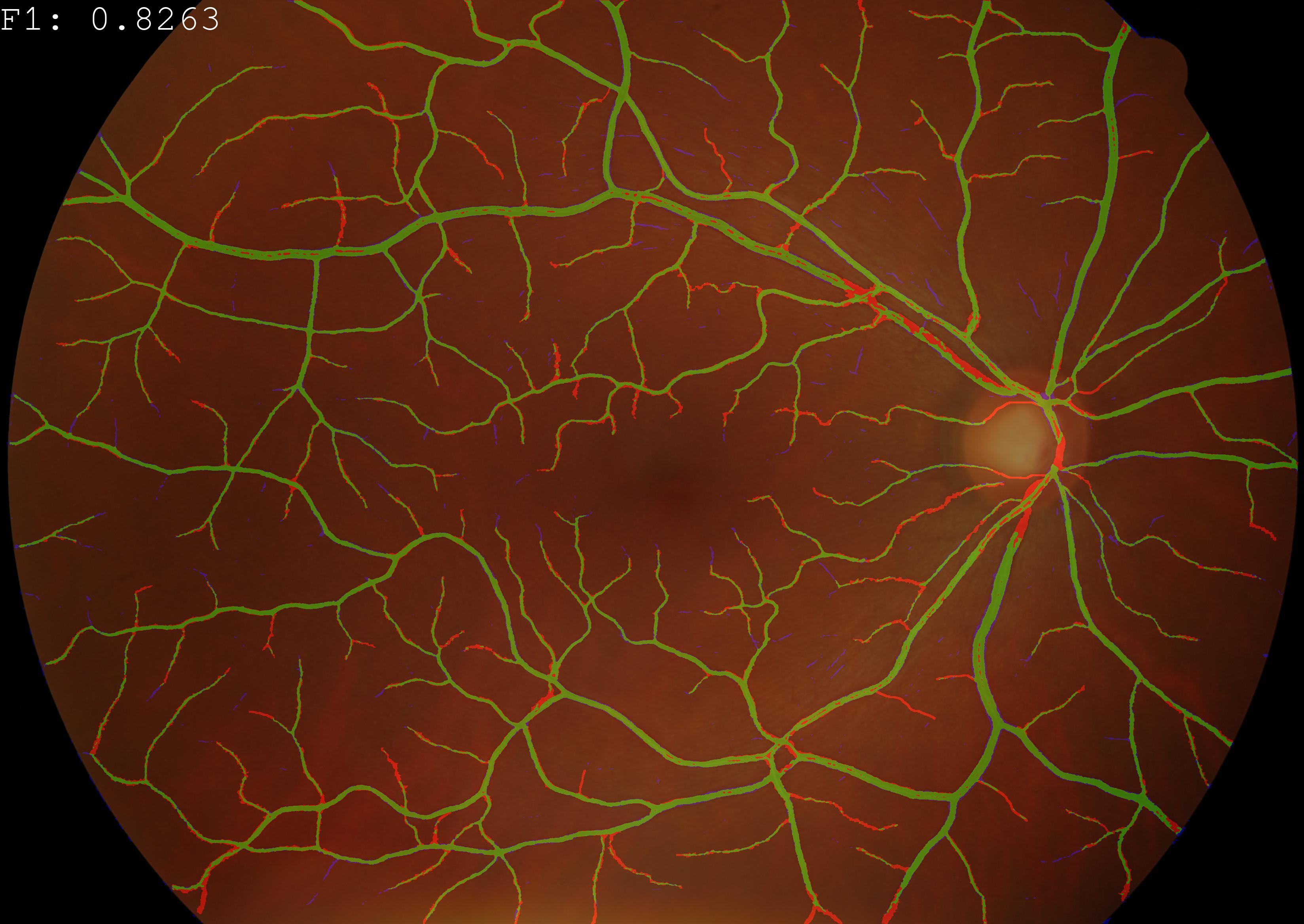}
    \end{minipage}
    \begin{minipage}[h]{0.32\textwidth}
        \includegraphics[width=\textwidth]{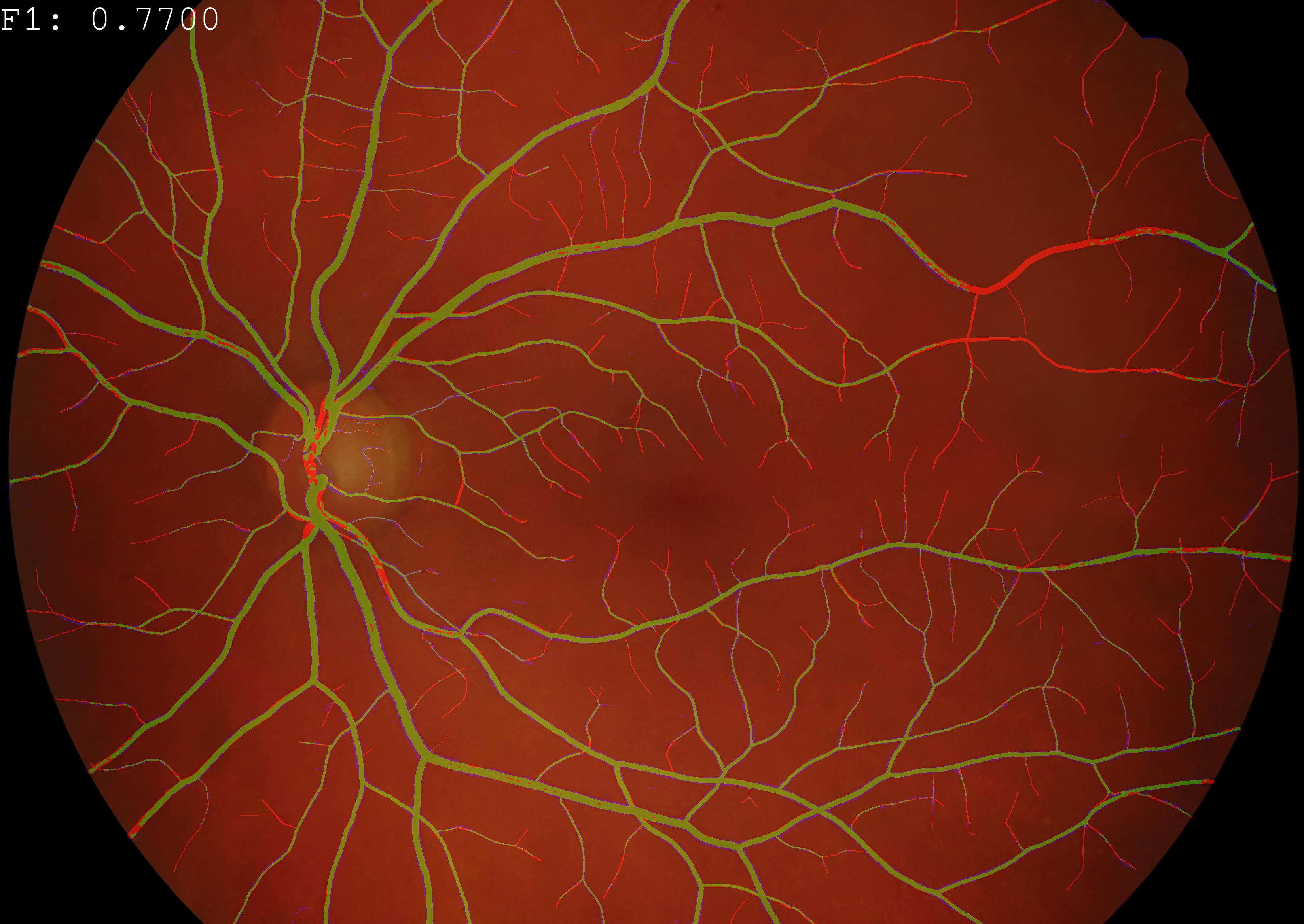}
    \end{minipage}
    \begin{minipage}[h]{0.32\textwidth}
        \includegraphics[width=\textwidth]{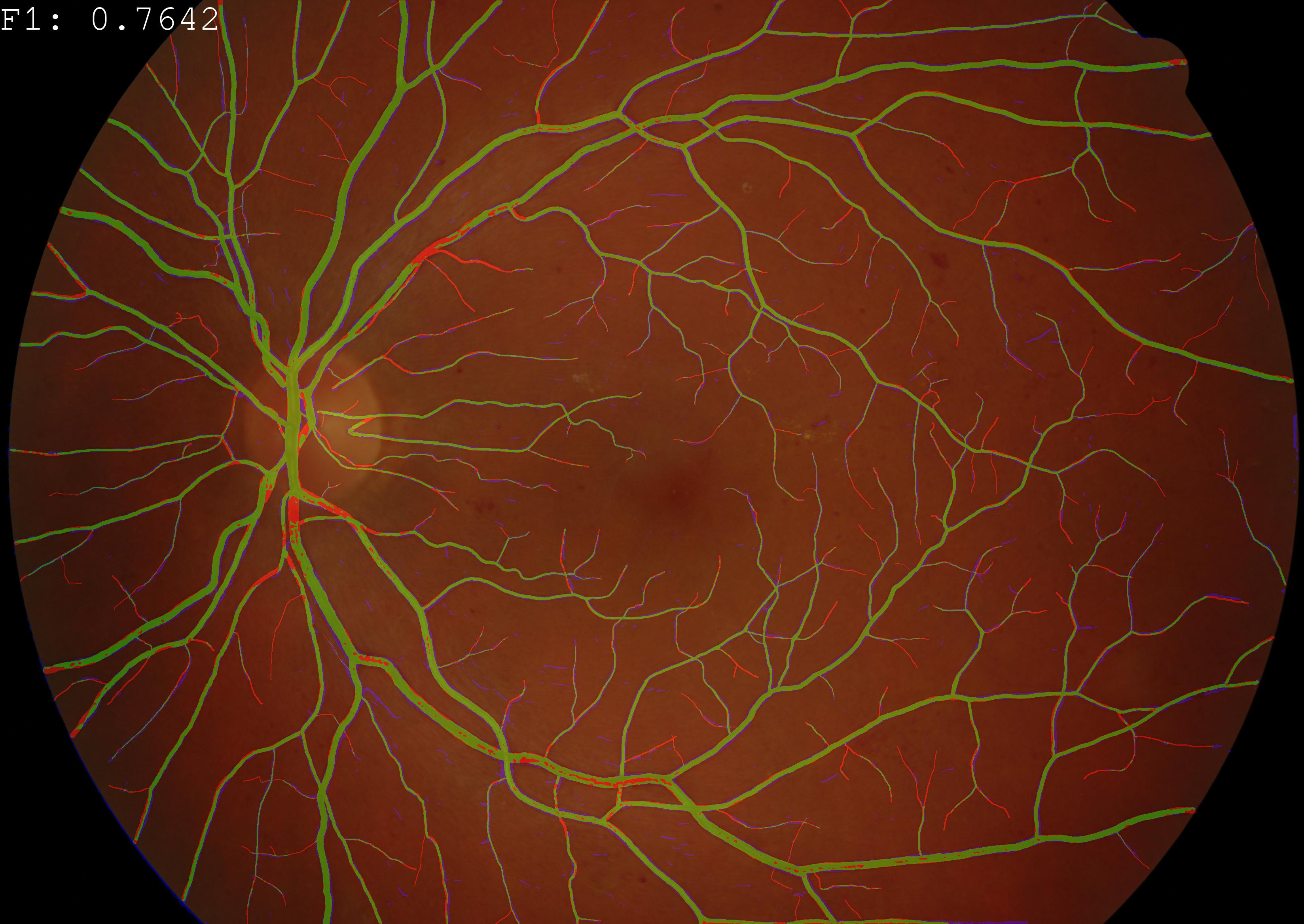}
    \end{minipage}
    \begin{minipage}[h]{0.02\textwidth}
    \rotatebox{90}{COVD}
    \end{minipage}
    \begin{minipage}[h]{0.32\textwidth}
        \includegraphics[width=\textwidth]{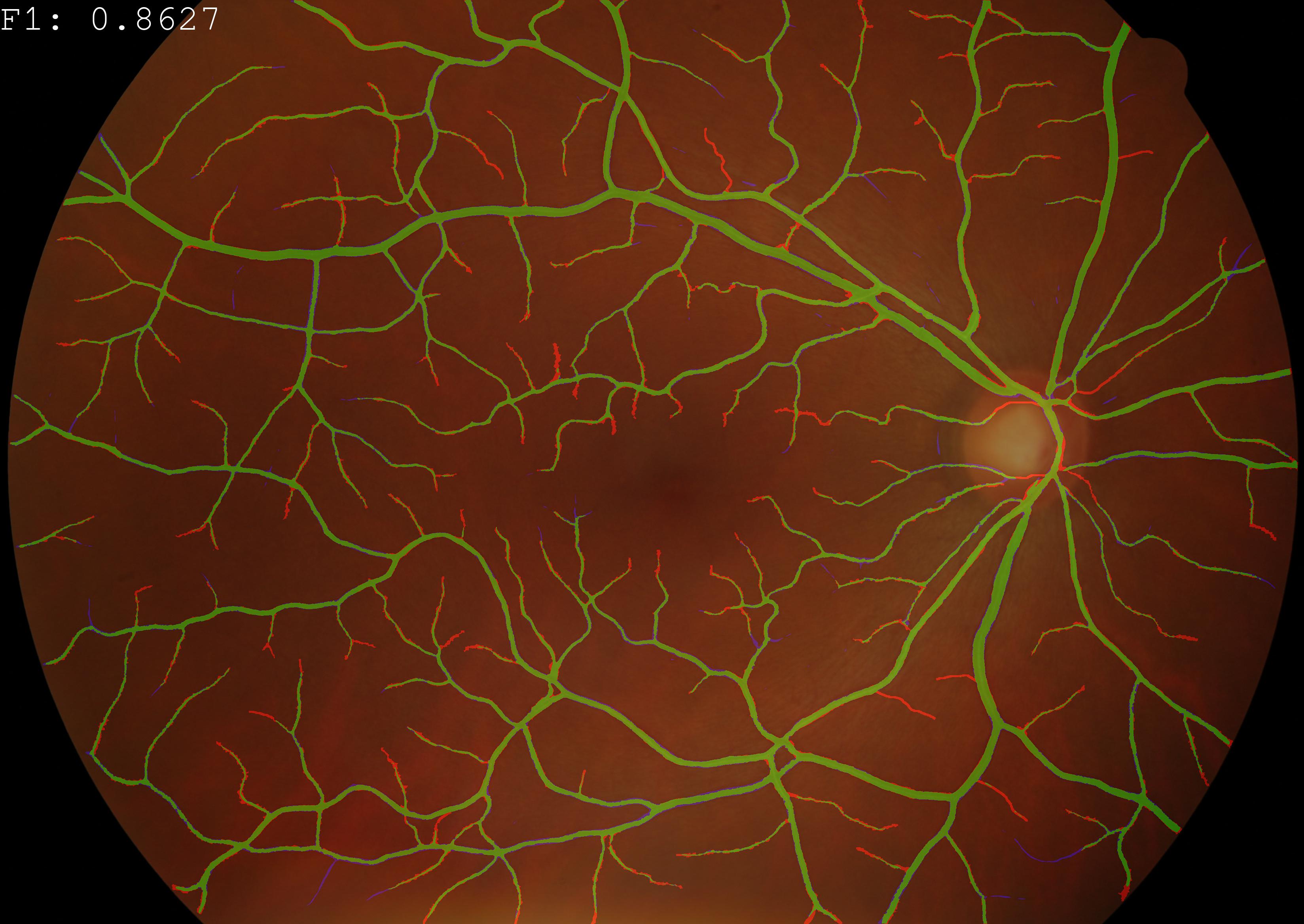}
    \end{minipage}
    \begin{minipage}[h]{0.32\textwidth}
        \includegraphics[width=\textwidth]{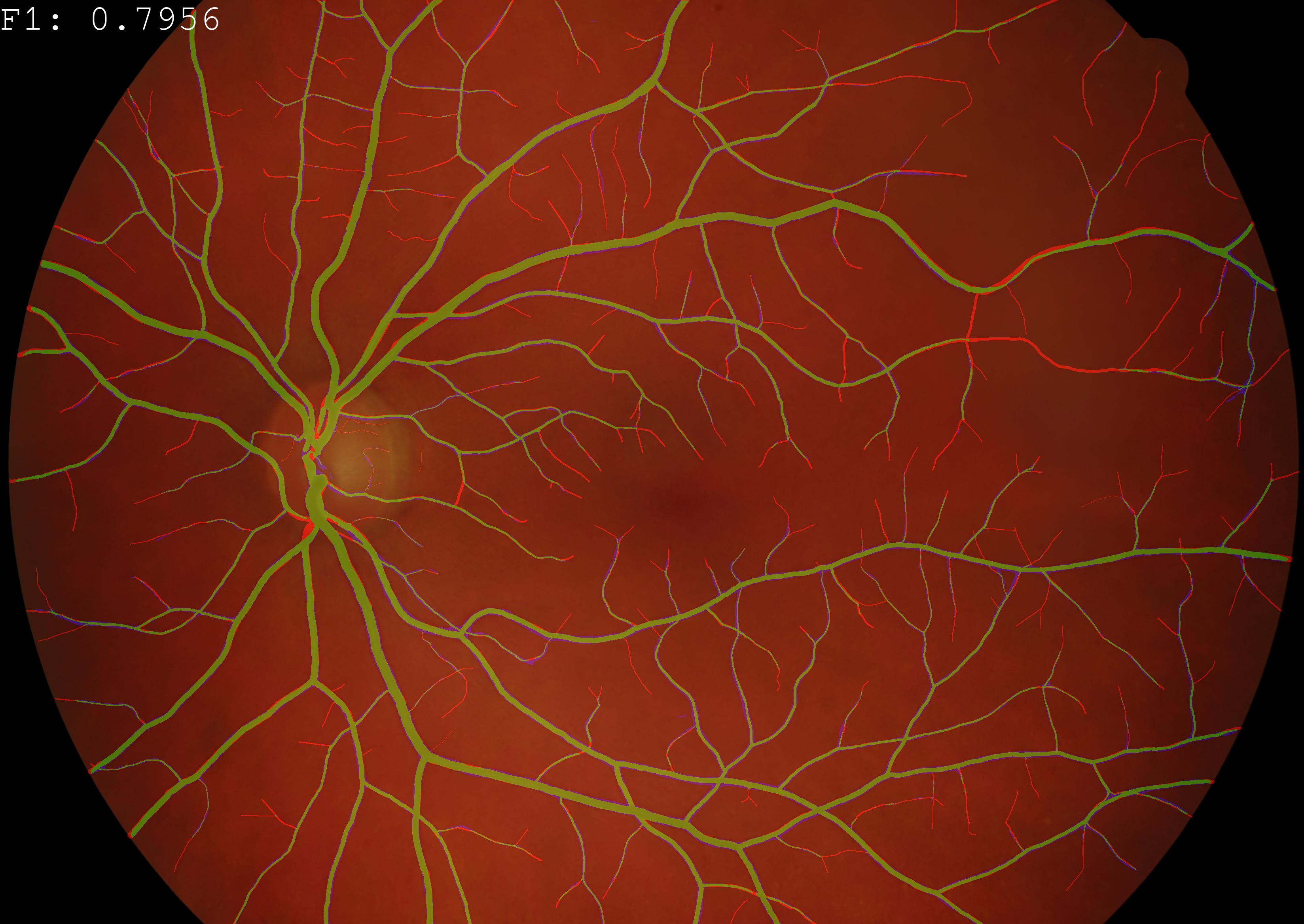}
    \end{minipage}
    \begin{minipage}[h]{0.32\textwidth}
        \includegraphics[width=\textwidth]{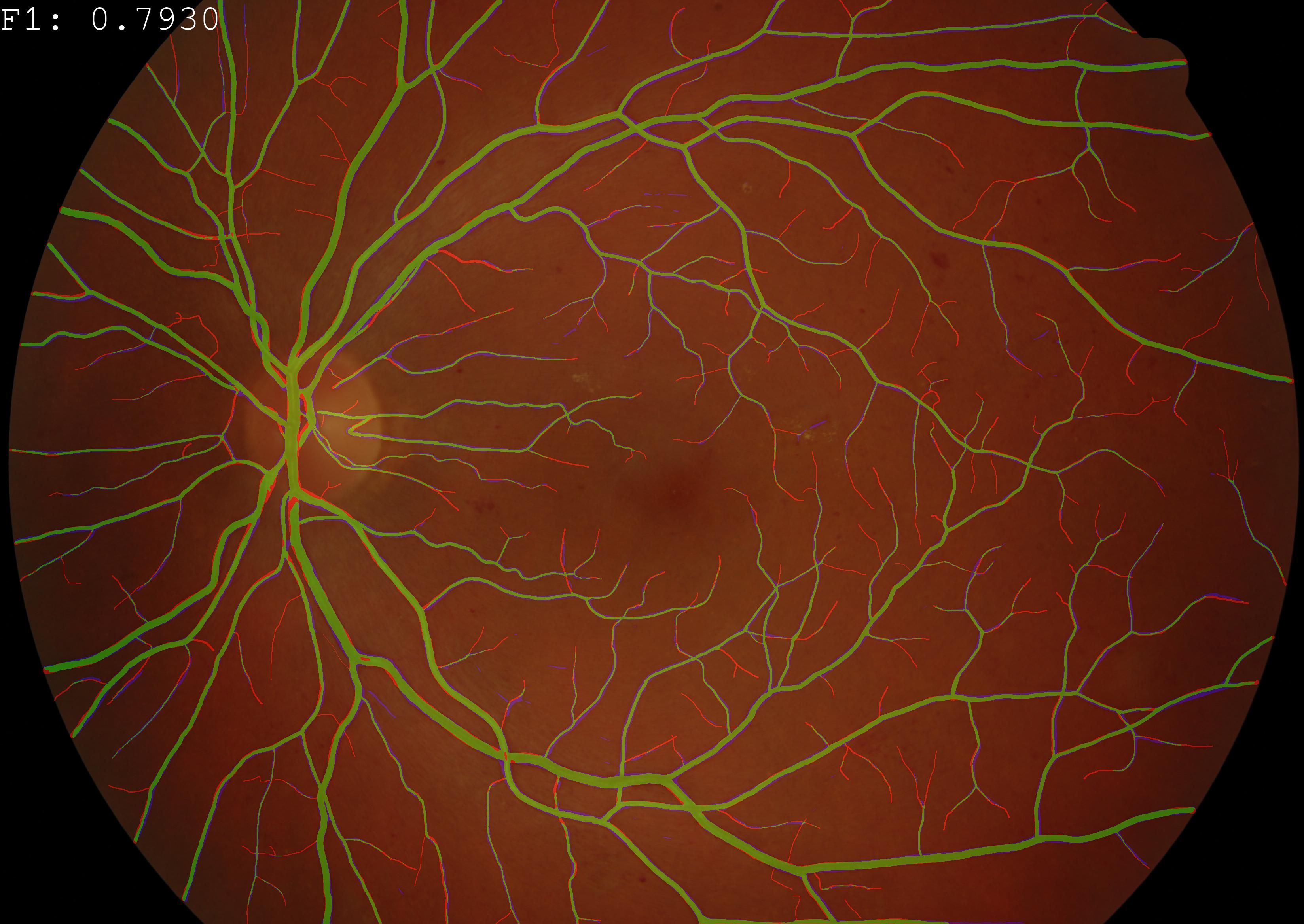}
    \end{minipage}
    \begin{minipage}[h]{0.02\textwidth}
    \rotatebox{90}{COVD-- SSL}
    \end{minipage}
    \begin{minipage}[h]{0.32\textwidth}
        \includegraphics[width=\textwidth]{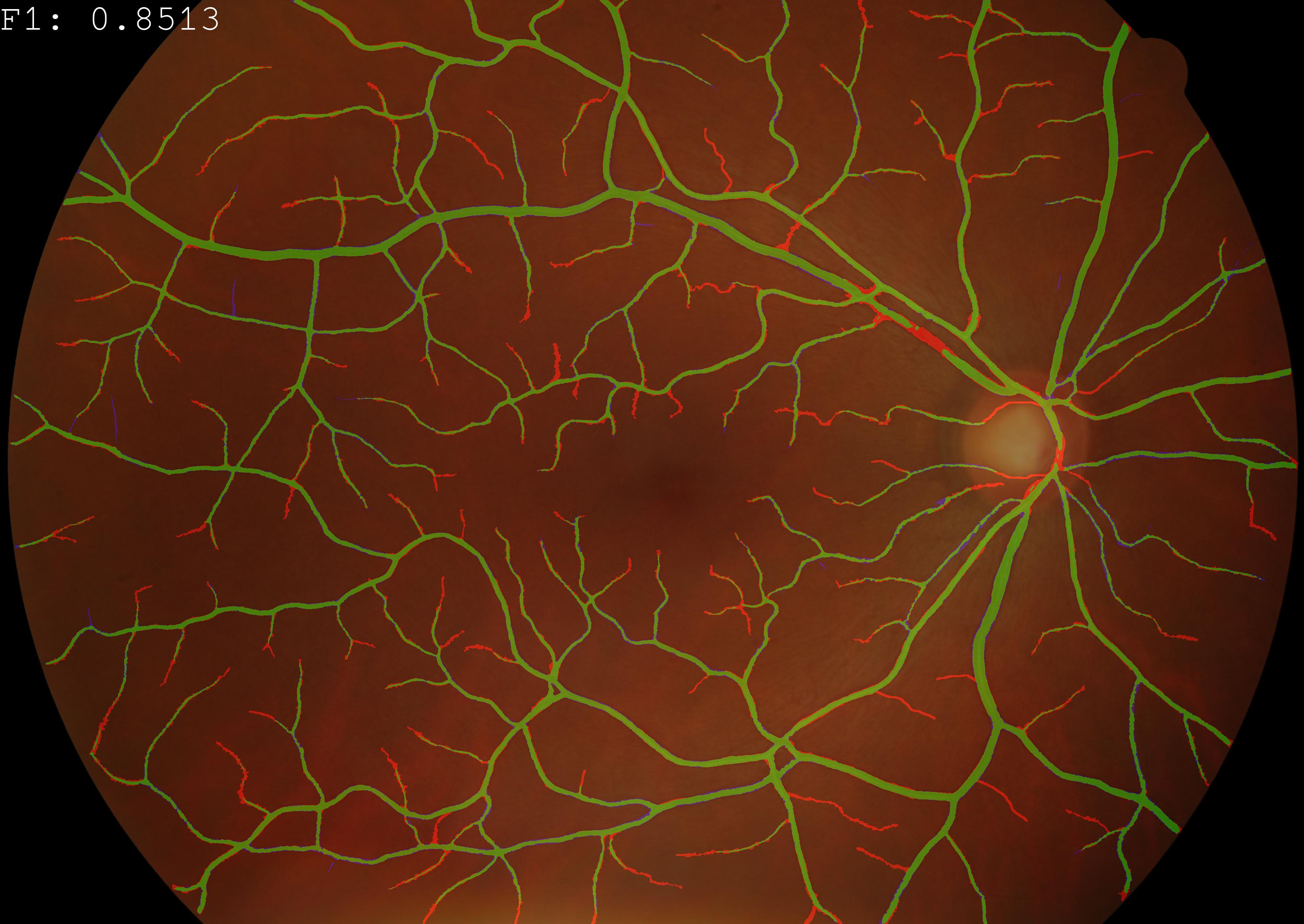}
    \end{minipage}
    \begin{minipage}[h]{0.32\textwidth}
        \includegraphics[width=\textwidth]{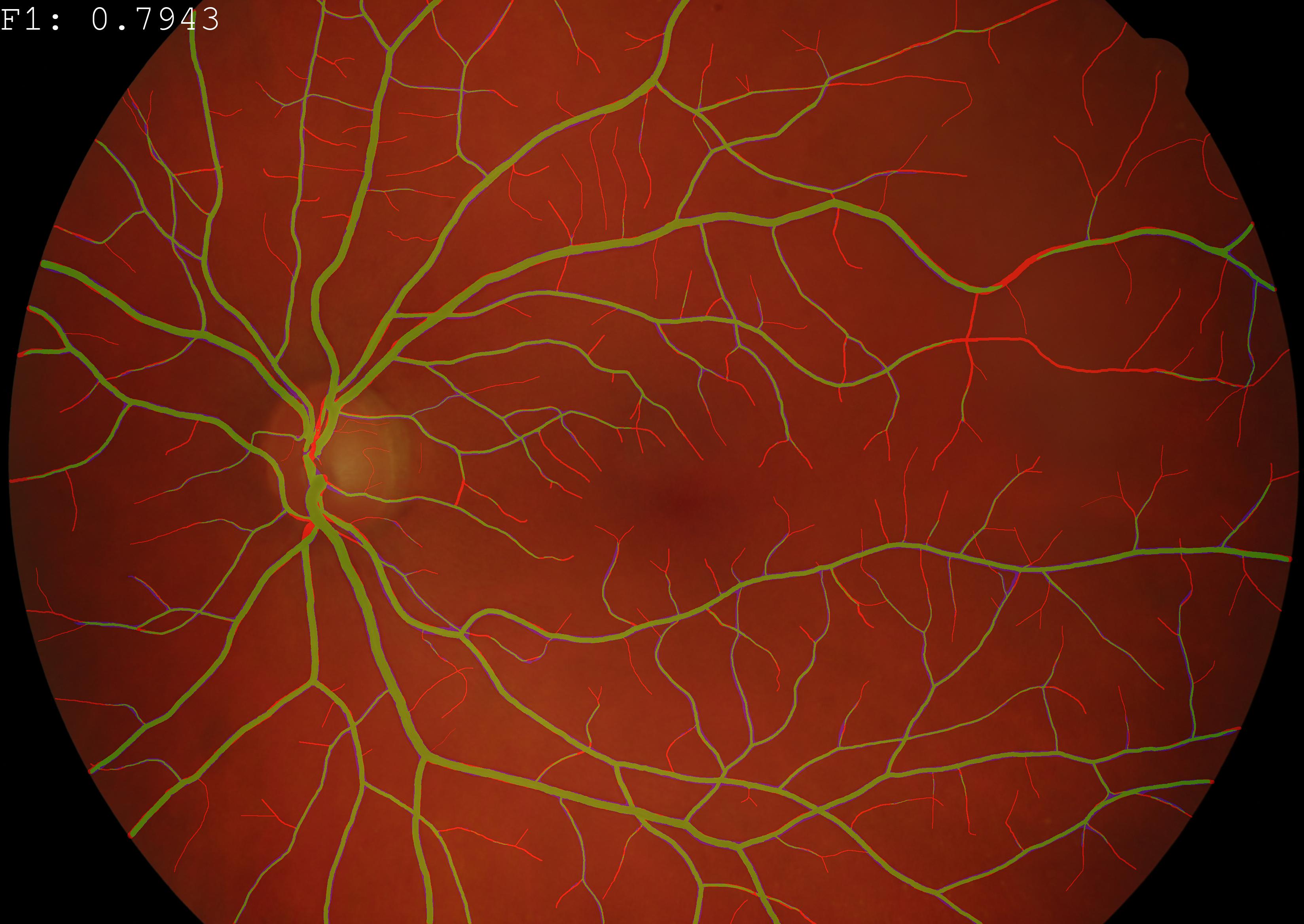}
    \end{minipage}
    \begin{minipage}[h]{0.32\textwidth}
        \includegraphics[width=\textwidth]{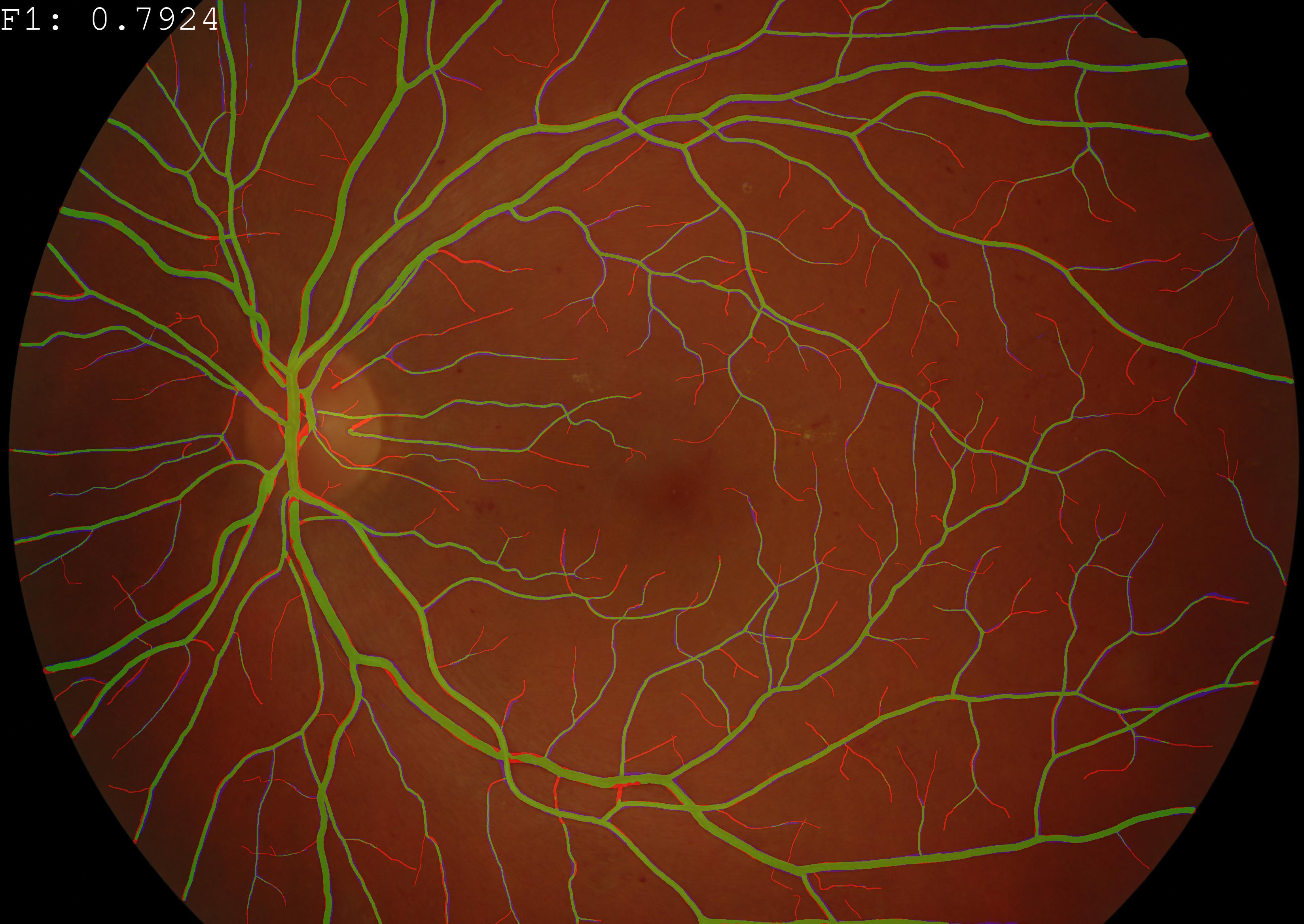}
    \end{minipage}
     \caption{Illustration of predicted vessel maps vs. ground truths for M2U-Net evaluated on the HRF test-set. The source dataset is indicated in the first column. True positives, false positives and false negatives are displayed in green, blue and red respectively.}
    \label{fig:hrfviz}
\end{figure*}

To put our results into perspective, Table~\ref{tab:resultsoverview} compares previous works to our methods.  We report the $F1_{micro}$-score for our results together with their standard deviation. Overall M2U-Net and DRIU trained on COVD$-$ are competitive, with the best performance on the high-resolution dataset HRF, where a new state-of-the-art F1-score could be reached. Additional metrics such as accuracy, sensitivity, precision and specificity are available in the documentation of our software package.  Further, to the available public datasets, we trained M2U-Net on COVD for a private target dataset with a resolution of 1920x1920 for which no ground-truth data is available. The predicted vessel probability maps are displayed in Figure \ref{fig:gnbtopcon}. While we can only make subjective statements, we find that the generated probability maps are of good quality, with the majority of main vessels being detected, even in presence of pathologies.  An illustration of predicted vessel maps \textit{versus} ground truths for M2U-Net on HRF is provided in Figure~\ref{fig:hrfviz}.
\begingroup
\setlength\tabcolsep{4pt} 

\begin{table*}[htb]
\scriptsize
\caption{Comparison with previous works on DRIVE. For our work we report $F1_{micro}$ scores together with their standard deviation}
\label{tab:resultsoverview}
\begin{center}
\begin{tabular}{@{}ccccccc@{}}
\toprule
                                                    &      & \multicolumn{5}{c}{F1-scores (std)}                                                               \\ \cmidrule(lr){3-7}
Method                                              & Year &  DRIVE (584 x 565)              & STARE (605 x 700)           & CHASE\_DB1 (960 x 999)                & IOSTAR (1024 x 1024)                   & HRF (2336 x 3504)                       \\ \hline
2nd human observer                                  &      & 0.7931*            & 0.752            & 0.7686                     & -                         & -                          \\
\textbf{Unsupervised}                               &      &                    &                  &                            &                           &                            \\       
Bibiloni et al.~\cite{bibiloni_real-time_2018}      & 2018 & 0.7521*            & -                & -                          & -                         & -                          \\      
\textbf{Supervised}                                 &      &                    &                  &                            &                           &                            \\ \hline     
Annunziata et al.~\cite{annunziata_leveraging_2016} & 2016 & -                  & 0.7683           & -                          & -                         & 0.7578                     \\  
Fraz et al.~\cite{fraz_ensemble_2012}               & 2012 & 0.7929*            & 0.7747           & 0.7566*                    & -                         & -                          \\
Jin et al.~\cite{jin_dunet_2019}                    & 2019 & \textbf{0.8237*}   & 0.8143           & 0.7883                     & -                         & 0.7988*                    \\ 
Laibacher et al.~\cite{laibacher_m2u_2018}          & 2018 & 0.8091*            & -                & 0.8006*                    & -                         & 0.7814                     \\ 
Maninis et al.~\cite{maninis_deep_2016}             & 2016 & 0.8220*            & \textbf{0.831}* & -                           & -                         & -                          \\   
Marin et al.~\cite{marin_new_2011}                  & 2011 & 0.8134*            & 0.8080           & -                          & -                         & -                          \\  
Orlando et al.~\cite{orlando_discriminatively_2017} & 2017 & 0.7857*            & 0.7644           & 0.7332                     & -                         & 0.7158*                    \\
Yan et al.~\cite{yan_joint_2018}                    & 2018 & 0.8183*            & -                & -                          & -                         & 0.7212*                    \\ 
Zhao et al.~\cite{zhao_supervised_2019}             & 2019 & 0.7882*            & 0.7960*          & -                          & 0.7707                    & 0.7659                     \\ 
\textbf{M2U-Net Target}                             & 2019 & 0.8030 (0.0142)    & 0.8150 (0.0411)  & \textbf{0.8022} (0.0193)   & \textbf{0.8173} (0.0203)  & 0.7800 (0.0574)            \\ \hline
\textbf{M2U-Net COVD --}                            & 2019 & 0.7885 (0.0179)    & 0.8117 (0.0456)  & 0.7884 (0.0238)            & 0.7928 (0.0153)           & \textbf{0.8020} (0.0445)   \\
\textbf{DRIU COVD --}                               & 2019 & 0.7877 (0.0182)    & 0.7775 (0.1171)  & 0.7964 (0.0269)            & 0.7910 (0.0206)           & 0.7994 (0.0439)            \\
\textbf{M2U-Net COVD -- SSL}                        & 2019 & 0.7913 (0.0143)    & 0.8196 (0.0443)  & 0.7988 (0.0260)            & 0.7845 (0.0181)           & 0.7972 (0.0436)            \\ \hline 
\multicolumn{7}{l}{*Same train-test split as adopted in this work}  \\ \bottomrule

\\ \bottomrule
\end{tabular}
\end{center}
\end{table*}
\endgroup

\section{Conclusions}
\label{sec:conclusion}
In this work we showed that simple transformation techniques like rescaling, padding and cropping of combined lower-resolution source datasets to the resolution and spatial composition of a higher-resolution target dataset can be a surprisingly effective way to improve segmentation quality in unseen conditions.  In comparison to previous works, we note that our approach achieves comparable performance without utilizing any training images of the target-set, solely relying on the remaining publicly available datasets for training. Given the stark differences in color, illumination, brightness and spatial composition between datasets this is an encouraging result and illustrates the robustness of our approach. Our experiments with semi-supervised learning were not conclusive.

We emphasized the need for a more rigorous and detailed focus on evaluation, and proposed a set of plots and metrics that give additional insights into model performance.  We propose to use the standard deviation as a proxy for the confidence on the estimation of the average F1-score.  We demonstrated via tables and plots how to take advantage of that information, throwing a new light over some published benchmarks.  We argue the performance of many contributions available in literature is actually quite comparable within standard deviation margins of each other, in spite of huge differences in the number of parameters for different architectures.  Lastly, we made our findings reproducible, distributing code and documentation for future researchers to build upon, in the hopes to inspire future work in the field.

\section*{Acknowledgments}
\begin{figure*}[!t]
\centering
    \begin{minipage}[h]{0.245\textwidth}
        \includegraphics[width=\textwidth]{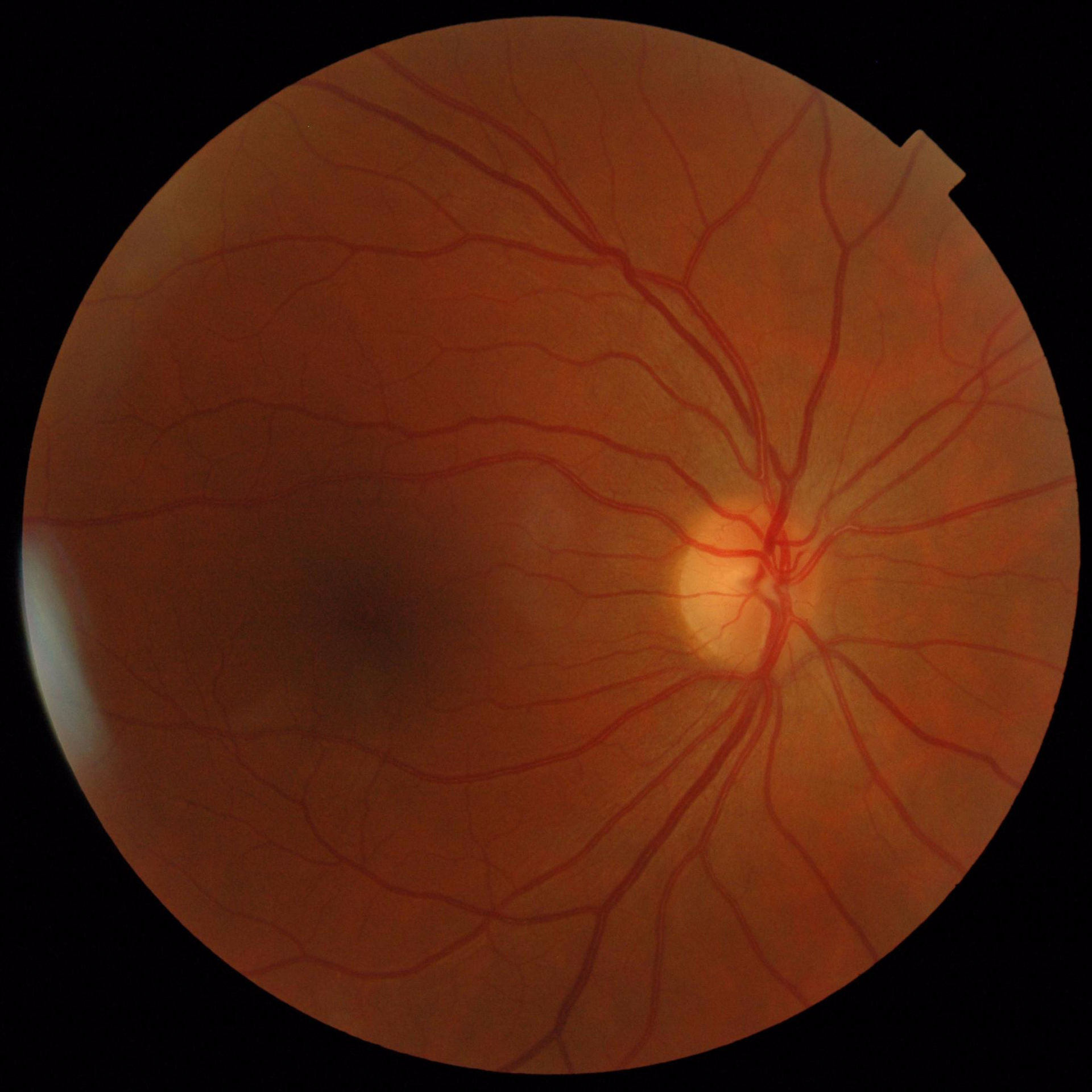}
    \end{minipage}
    \begin{minipage}[h]{0.245\textwidth}
        \includegraphics[width=\textwidth]{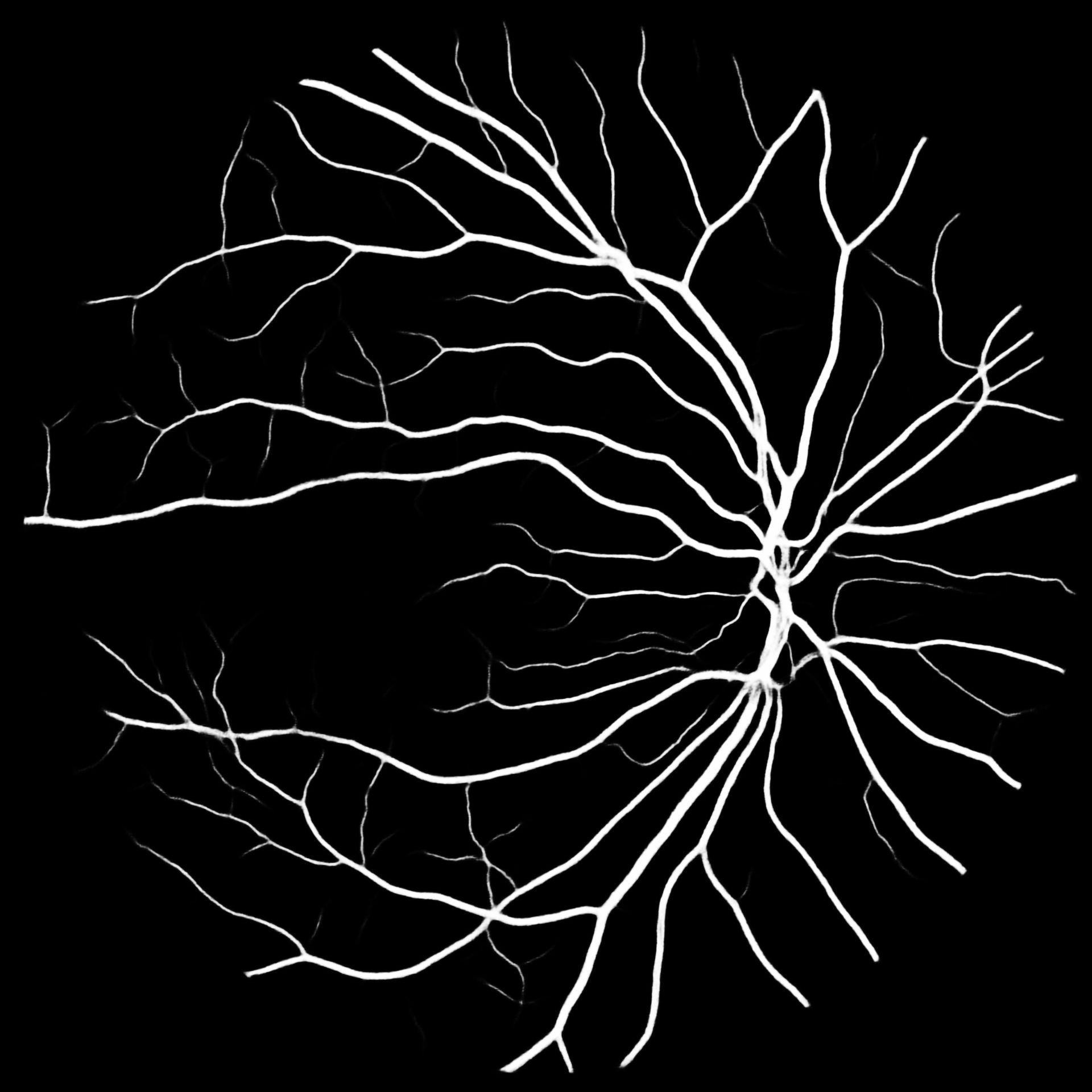}
    \end{minipage}
    \begin{minipage}[h]{0.245\textwidth}
        \includegraphics[width=\textwidth]{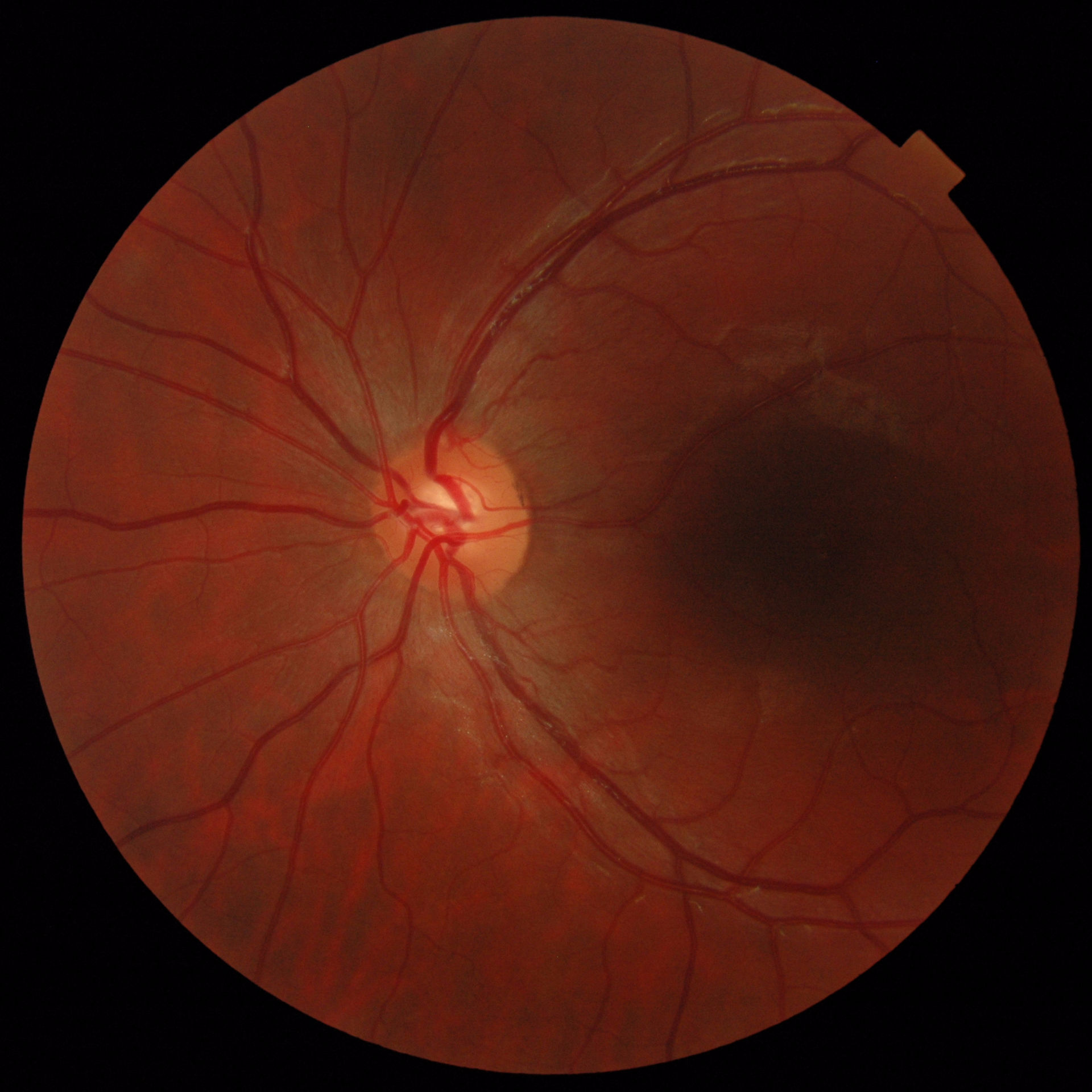}
    \end{minipage}
    \begin{minipage}[h]{0.245\textwidth}
        \includegraphics[width=\textwidth]{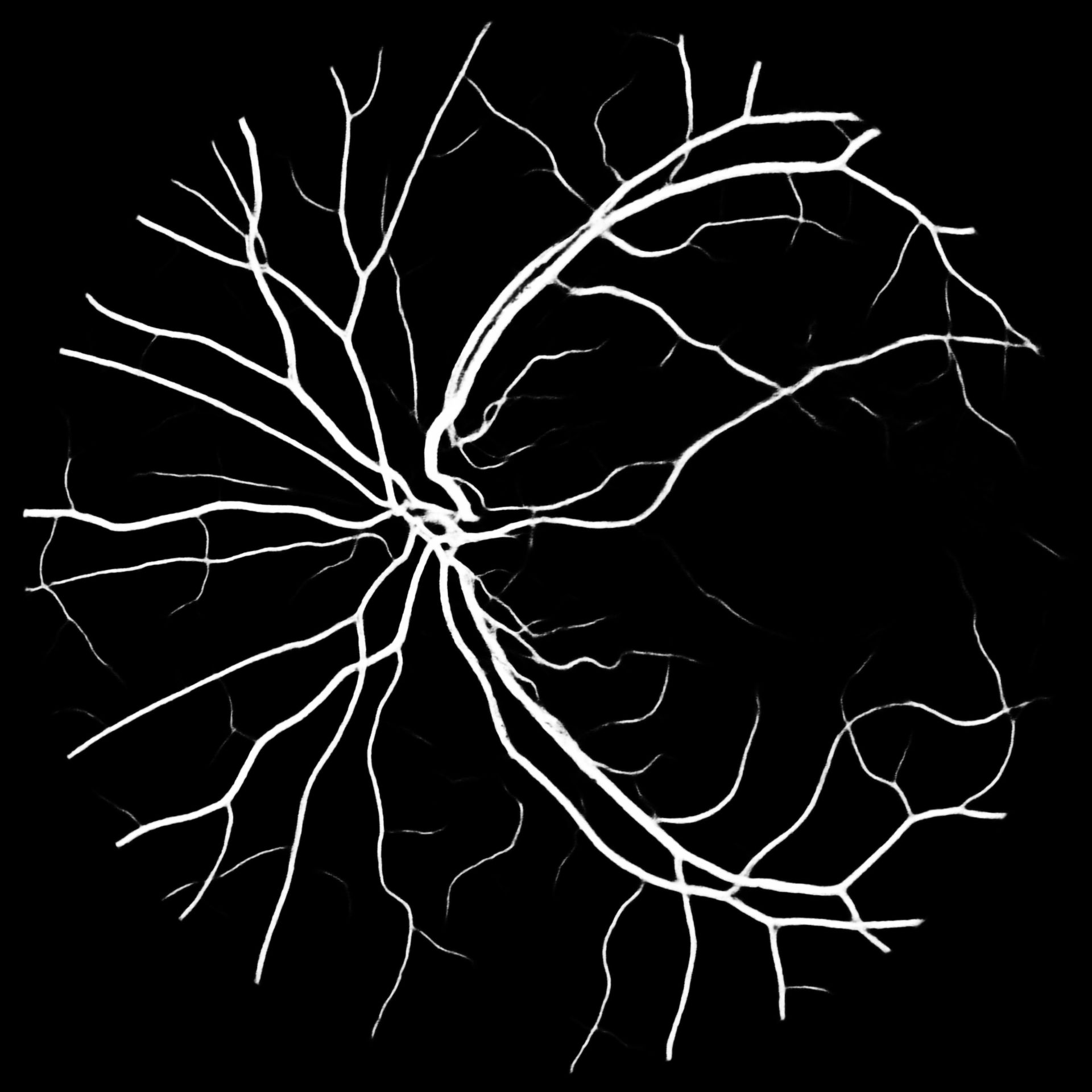}
    \end{minipage}
    
    \begin{minipage}[h]{0.245\textwidth}
        \includegraphics[width=\textwidth]{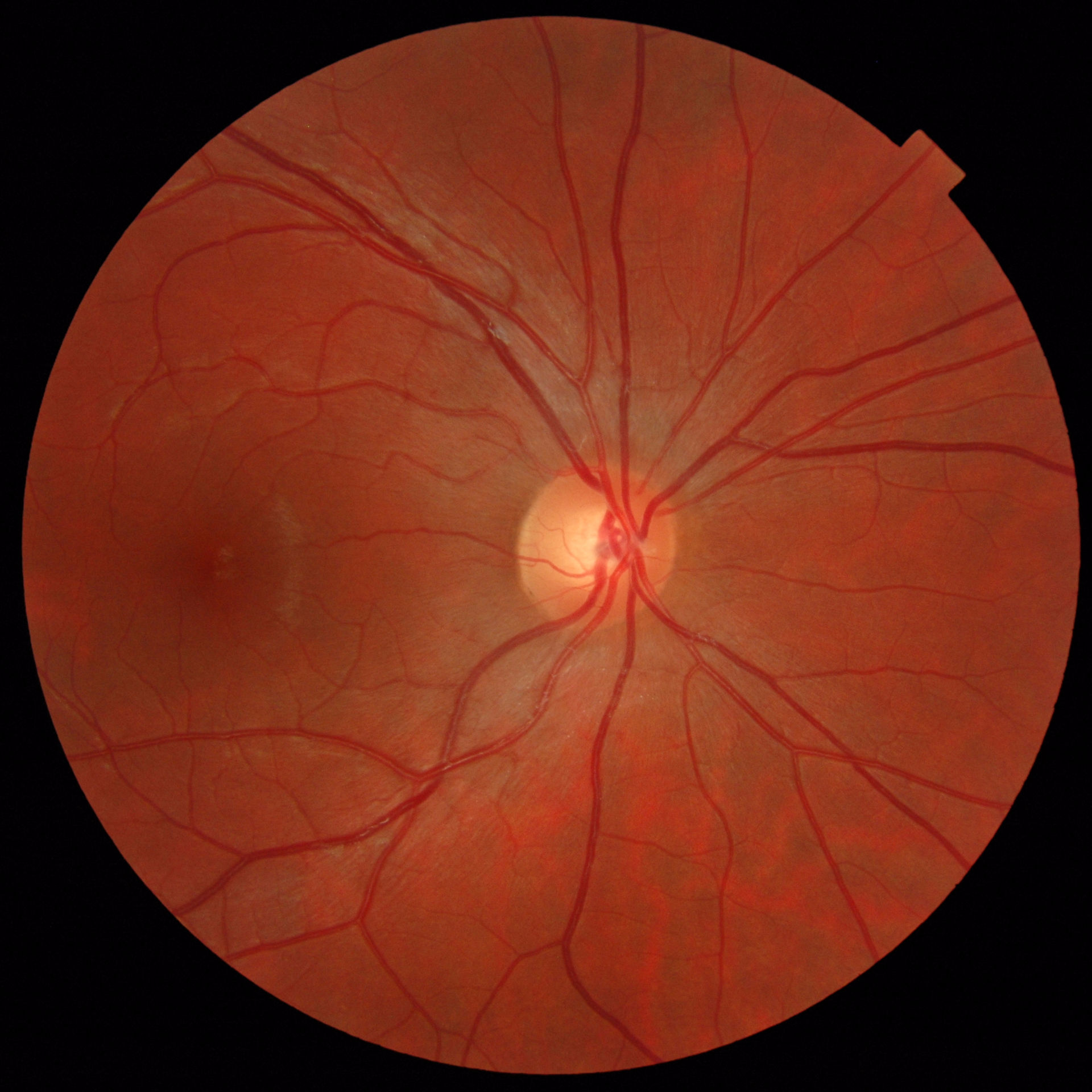}
    \end{minipage}
    \begin{minipage}[h]{0.245\textwidth}
        \includegraphics[width=\textwidth]{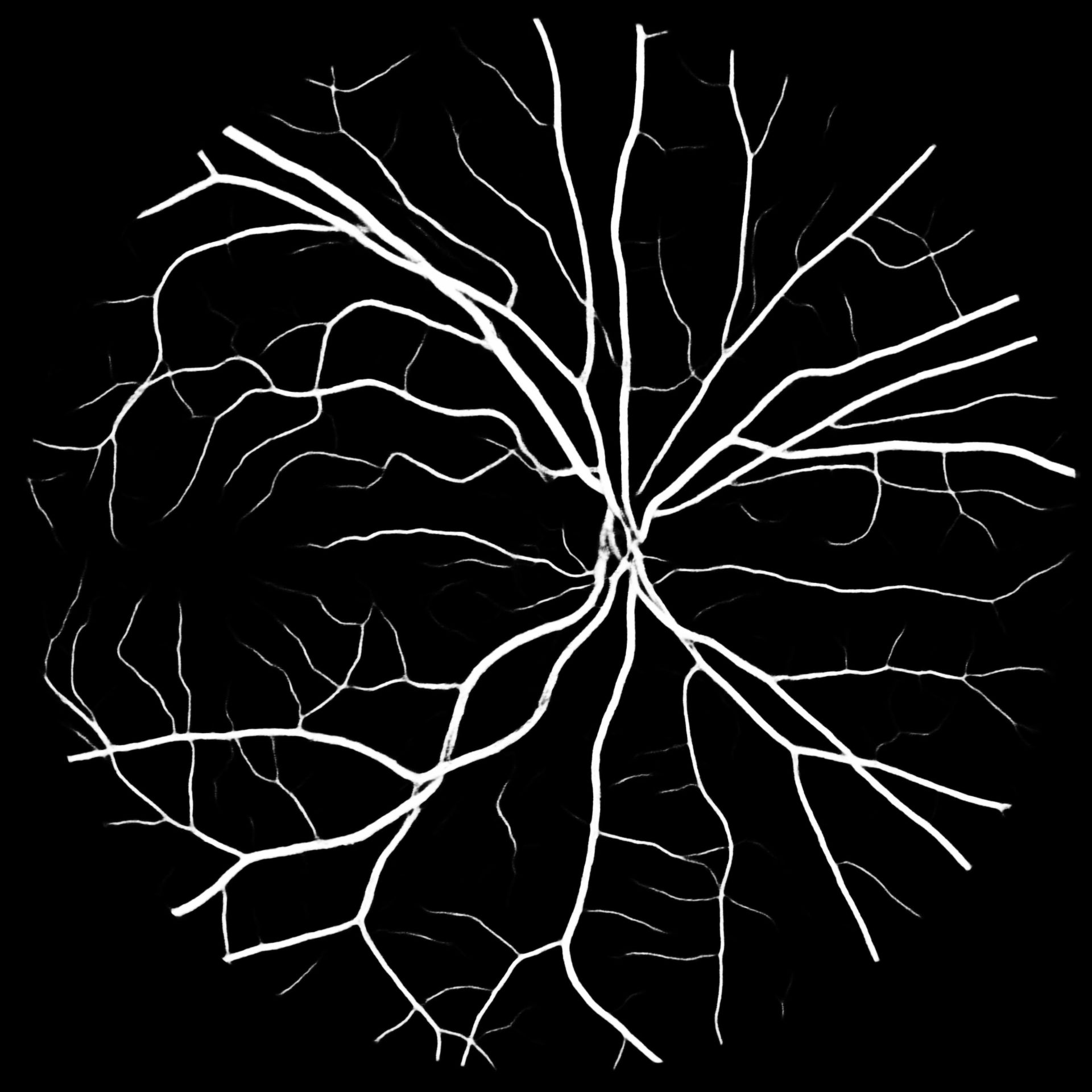}
    \end{minipage}
    \begin{minipage}[h]{0.245\textwidth}
        \includegraphics[width=\textwidth]{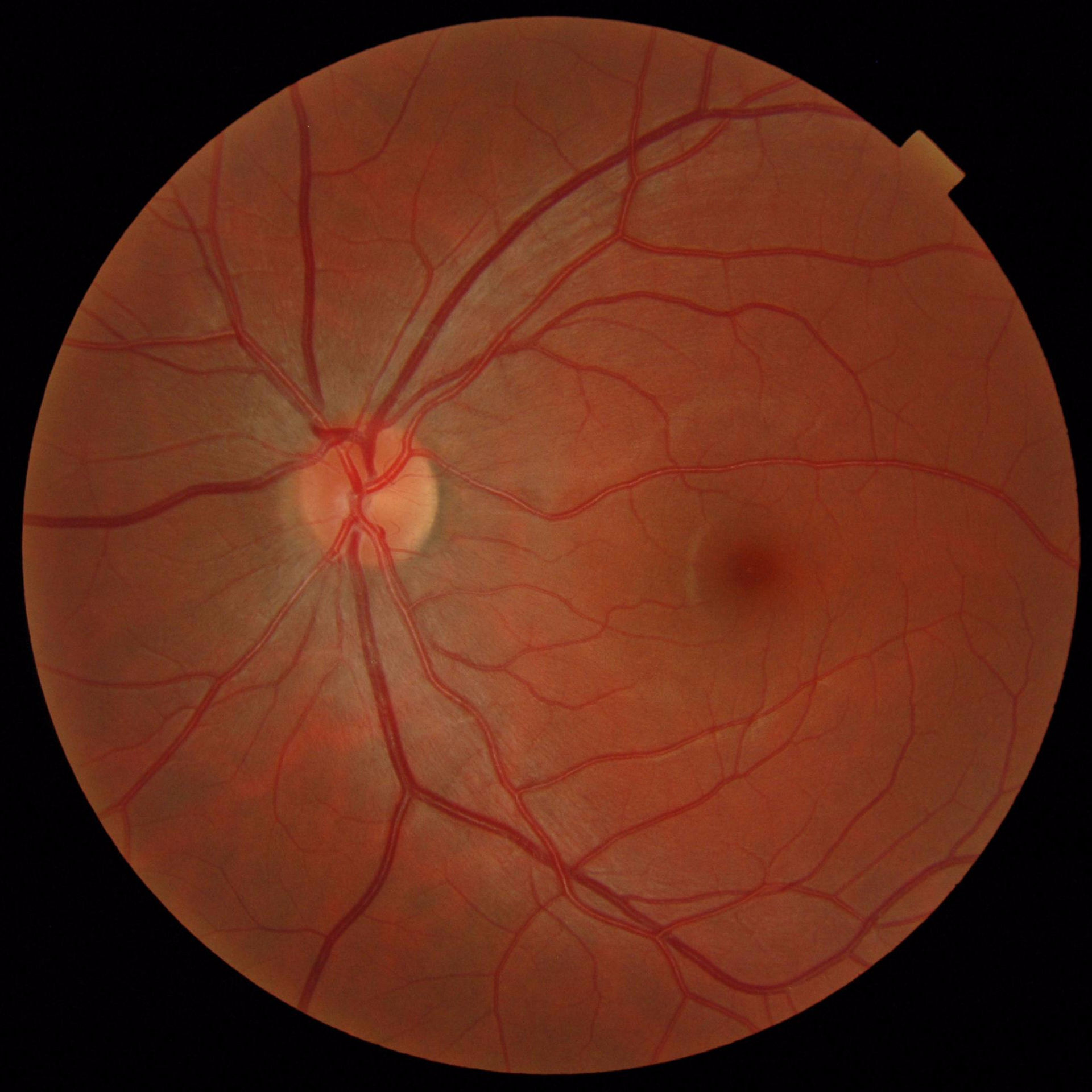}
    \end{minipage}
    \begin{minipage}[h]{0.245\textwidth}
        \includegraphics[width=\textwidth]{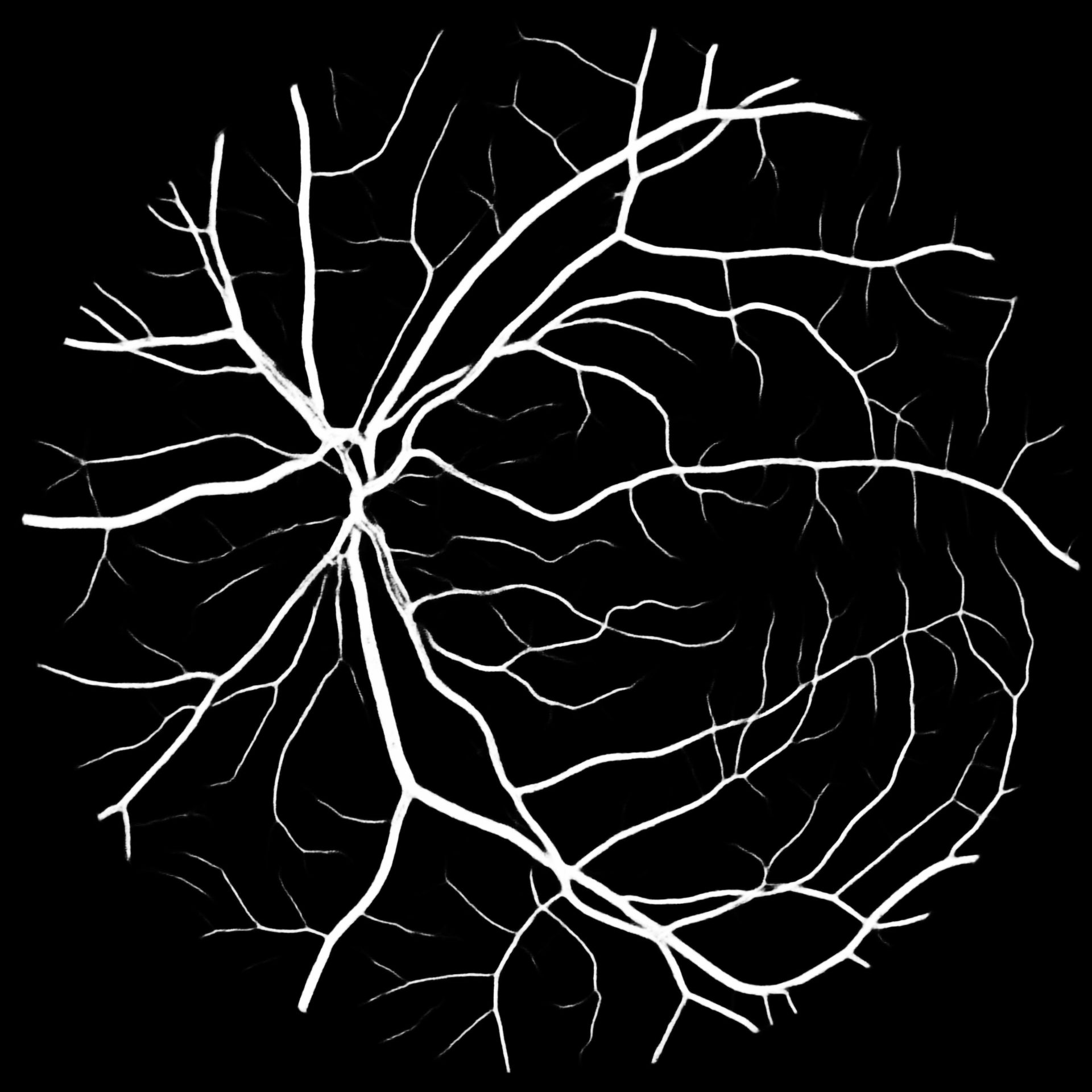}
    \end{minipage}
    
     \caption{Illustration of predicted vessel maps for M2U-Net, trained on COVD, applied on a private high-resolution dataset (1920x1920) for which no ground-truth data is available.}
    \label{fig:gnbtopcon}
\end{figure*}

The authors would like to thank Dr. Christophe Chiquet from the University Hospital of Grenoble and Inserm, France, for providing the high-resolution retina fundus images in Fig.~\ref{fig:gnbtopcon}.

\bibliographystyle{IEEEtran}
\bibliography{bibtex/bib/binseg.bib}
\end{document}